\documentclass[times,twocolumn,final]{elsarticle}

\usepackage{medima}
\usepackage{framed,multirow}

\usepackage{amssymb}
\usepackage{latexsym}
\usepackage{listings}
\definecolor{mygreen}{rgb}{0,0.6,0}
\definecolor{mymauve}{rgb}{0.58,0,0.82}

\usepackage[linesnumbered,ruled,vlined]{algorithm2e}

\SetAlgoSkip{0.5em}                
\usepackage{amsmath}
\usepackage{diagbox}
\usepackage[flushleft]{threeparttable}
\usepackage{pifont}
\newcommand{\cmark}{\ding{51}}%
\newcommand{\xmark}{\ding{55}}%
\usepackage[caption=false, font=normalsize, labelfont=sf, textfont=sf]{subfig}
\usepackage{array}
\usepackage{booktabs}
\usepackage[dvipsnames]{xcolor}

\usepackage{hyperref}

\definecolor{newcolor}{rgb}{.8,.349,.1}

\journal{Artificial Intelligence In Medicine}

\begin{document}

\verso{Zhan Xiong \textit{et~al.}}

\begin{frontmatter}

\title{Advances in Kidney Biopsy Lesion Assessment through Dense Instance Segmentation}%

\author[1]{Zhan \snm{Xiong}}
\author[2]{Junling \snm{He}}
\author[3]{Pieter \snm{Valkema}}
\author[4]{Tri Q. \snm{Nguyen}}
\author[5,6]{Maarten \snm{Naesens}}
\author[2,3,7,8]{Jesper \snm{Kers}\corref{cor2}\fnref{fn1}}
\fntext[fn1]{Data and annotations can be made available upon reasonable request due to the cooperation with multiple hospitals. Please email to J. Kers, MD, PhD: j.kers@amsterdamumc.nl.
Source code is available on request from LIACS gitlab: f.j.verbeek@liacs.leidenuniv.nl}

\author[1,8]{Fons J. \snm{Verbeek}\corref{cor1}}
\cortext[cor1]{Corresponding author.
Email address: f.j.verbeek@liacs.leidenuniv.nl}

\address[1]{LIACS, Leiden University, Snellius Gebouw, Niels Bohrweg 1, 2333 CA, Leiden, The Netherlands}
\address[2]{Department of Pathology, Leiden University Medical Center, Albinusdreef 2, 2333 ZA, Leiden, The Netherlands}
\address[3]{Department of Pathology, Amsterdam UMC, University of Amsterdam, Meibergdreef 9, 1105 AZ, Amsterdam, The Netherlands}
\address[4]{Department of Pathology, University Medical Center Utrecht, Heidelberglaan 100, 3584 CX, Utrecht, The Netherlands}
\address[5]{Department of Nephrology and Renal Transplantation, University Hospitals Leuven, Herestraat 49, 3000, Leuven, Belgium}
\address[6]{Department of Microbiology, Immunology, and Transplantation, KU Leuven, Oude Markt 13, 3000, Leuven, Belgium}
\address[7]{Van't Hoff Institute for Molecular Sciences, University of Amsterdam, Science Park 904, 1098 XH, Amsterdam, The Netherlands}
\address[8]{Shared senior authorship}

\received{}
\finalform{}
\accepted{}
\availableonline{}
\communicated{}

\begin{abstract}
Renal biopsies are the gold standard for the diagnosis of kidney diseases. Lesion scores made by renal pathologists are semi-quantitative and exhibit high inter-observer variability. Automating lesion classification within segmented anatomical structures can provide decision support in quantification analysis, thereby reducing inter-observer variability. Nevertheless, classifying lesions in regions-of-interest (ROIs) is clinically challenging due to (a) a large amount of densely packed anatomical objects, (b) class imbalance across different compartments (at least 3), (c) significant variation in size and shape of anatomical objects and (d) the presence of multi-label lesions per anatomical structure. Existing models cannot address these complexities in an efficient and generic manner. This paper presents an analysis for a \textbf{generalized solution} to datasets from various sources (pathology departments) with different types of lesions. Our approach utilizes two sub-networks: dense instance segmentation and lesion classification. We introduce \textbf{DiffRegFormer}, an end-to-end dense instance segmentation sub-network designed for multi-class, multi-scale objects within ROIs. Combining diffusion models, transformers, and RCNNs, DiffRegFormer {is a computational-friendly framework that can efficiently recognize over 500 objects across three anatomical classes, i.e., glomeruli, tubuli, and arteries, within ROIs.} In a dataset of 303 ROIs from 148 Jones' silver-stained renal Whole Slide Images (WSIs), our approach outperforms previous methods, achieving an Average Precision of 52.1\% (detection) and 46.8\% (segmentation). Moreover, our lesion classification sub-network achieves 89.2\% precision and 64.6\% recall on 21889 object patches out of the 303 ROIs. Lastly, our model demonstrates direct domain transfer to PAS-stained renal WSIs without fine-tuning.
\end{abstract}


\end{frontmatter}


\section{Introduction}
\label{sec:introduction}
\begin{figure}[!htbp]
	\centering
	\includegraphics[width=0.5\textwidth, keepaspectratio=true]{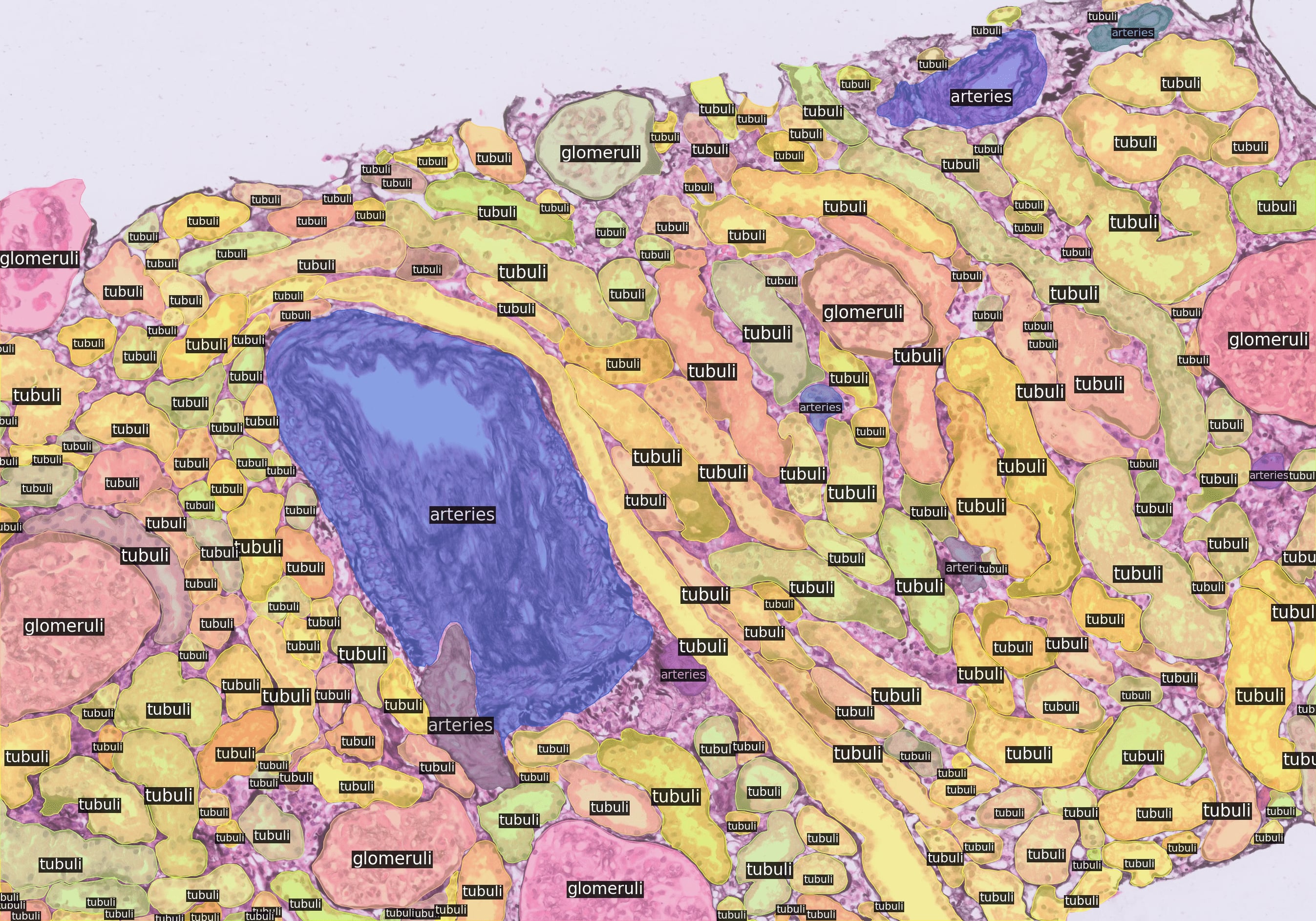}
	\caption{An illustration depicts a manually annotated ROI of kidney biopsy, focusing on \textbf{glomeruli, tubuli, and arteries}. There are three primary challenges encountered in clinical renal biopsies: (1) a large number of objects closely touching each other; (2) the significant variation in size and shape among different instances; (3) the distribution of classes is heavily biased.}
	\label{fig:illustration_gt_images}
\end{figure}

The evaluation of expert pathologists of kidney biopsies remains the gold standard for diagnosing and staging renal diseases \citep{brachemi2014renal}. Although biopsies digitalized into Whole Slide Images (WSIs)\footnote{https://www.mbfbioscience.com/whole-slide-imaging-analysis/} have facilitated obtaining a visual morphological assessment of different anatomical structures for disease categorization, high-quality diagnostic assessments heavily depend on the correct lesion quantification manually annotated by pathologists across structures within a biopsy. Figure \ref{fig:illustration_gt_images} shows an example of annotated region-of-interest (ROI)\footnote{To prevent any confusion with 'RoI' in RCNNs, we will hereafter use the term 'ROI' solely in the context of renal biopsies. To maintain clear distinctions in RCNN discussions, we will adopt the terms 'candidate regions' or 'regional proposals' s.} within a biopsy that contains hundreds of densely packed tissue objects. The annotation would cost a skilled expert around 2-4 hours for a complete biopsy. Due to the complexity and time-consuming nature of this task, there is a strong need for automated structure annotation and tools for lesion classification to facilitate further quantification, offload annotation time, and reduce intra- / inter-observer variability \citep{alnazer2021recent, srinidhi2021deep}. \par
Deep learning based instance segmentation algorithms have demonstrated significant capabilities on biomedical datasets \citep{xu2014deep, bouteldja2021deep, jiang2021deep,salvi2021automated,deng2023omni, yuan2023devil, lin2023gclr, meseguer2024micil, feng2024artificial}. However, developing a generic framework in renal pathology is still a challenge. This challenge can be elaborated in four technical gaps {and two practical issues.} The four gaps ({see} Figure \ref{fig:illustration_gt_images}) are: (1) densely packed structures, up to 1000, per ROI; (2) considerable variation in size and shape of objects (e.g., arteries can be up to 100 times larger than that of tubuli); (3) class imbalance (e.g., the tubulointerstitial area occupies more than 70\% on average in healthy and diseased renal parenchyma \citep{bouteldja2021deep}); {(4) each anatomical structure may present multiple lesions.} {The two practical issues are: (a) how to fuse multiple datasets with variation in staining to fully exploit scarce annotations; (b) readiness for extensibility, i.e., cost-effective adaptation to new lesion types from expanding datasets in clinical scenarios. Simultaneously addressing these difficulties requires a universal framework that includes efficiency, staining style (domain) transfer \citep{gadermayr2019generative, vasiljevic2021towards, kang2021stainnet}, and flexibility for continuous learning \citep{zhang2023continual, deng2024prpseg}}. \par
Prior studies have shown significant capabilities in lesion classification through dense instance segmentation, but their paradigms lack scalability and adaptability to datasets with potential changes in lesion compositions. Some approaches \citep{hermsen2019deep, bouteldja2021deep, salvi2021automated, liu2023diagnosis, deng2023omni} adopted a two-step process with semantic segmentation followed by dataset-specific post-processing to achieve final instance masks, which cannot be scaled to large-scale datasets with dense objects in various shapes. Besides, each lesion is defined as one semantic class, leading to difficulty for {multi-label lesion classification} and potential changes in lesion combinations. Specifically, adding or removing lesions requires complete redesign and retraining of the model. Lastly, increasing lesion types requires adding more segmentation maps for prediction, which is significantly resource-intensive. \par
Detection-based models with regional convolution neural networks (RCNNs) \citep{jha2021instance, jiang2021deep} can efficiently detect dense anatomical structures and separately design lesion classification heads for each class. They can, therefore, adopt a plug-and-play mechanism and adapt to lesion changes by replacing the corresponding sub-modules and maximally reusing the others. However, these variants are unscalable to multi-class objects with various scales and shapes due to the reliance on pre-defined bounding box anchors, limiting their application to process a single class with lesser variation in scales and shapes, e.g., glomeruli. \par
\begin{figure}[!htbp]
	\centering
	\includegraphics[width=0.5\textwidth, keepaspectratio=true]{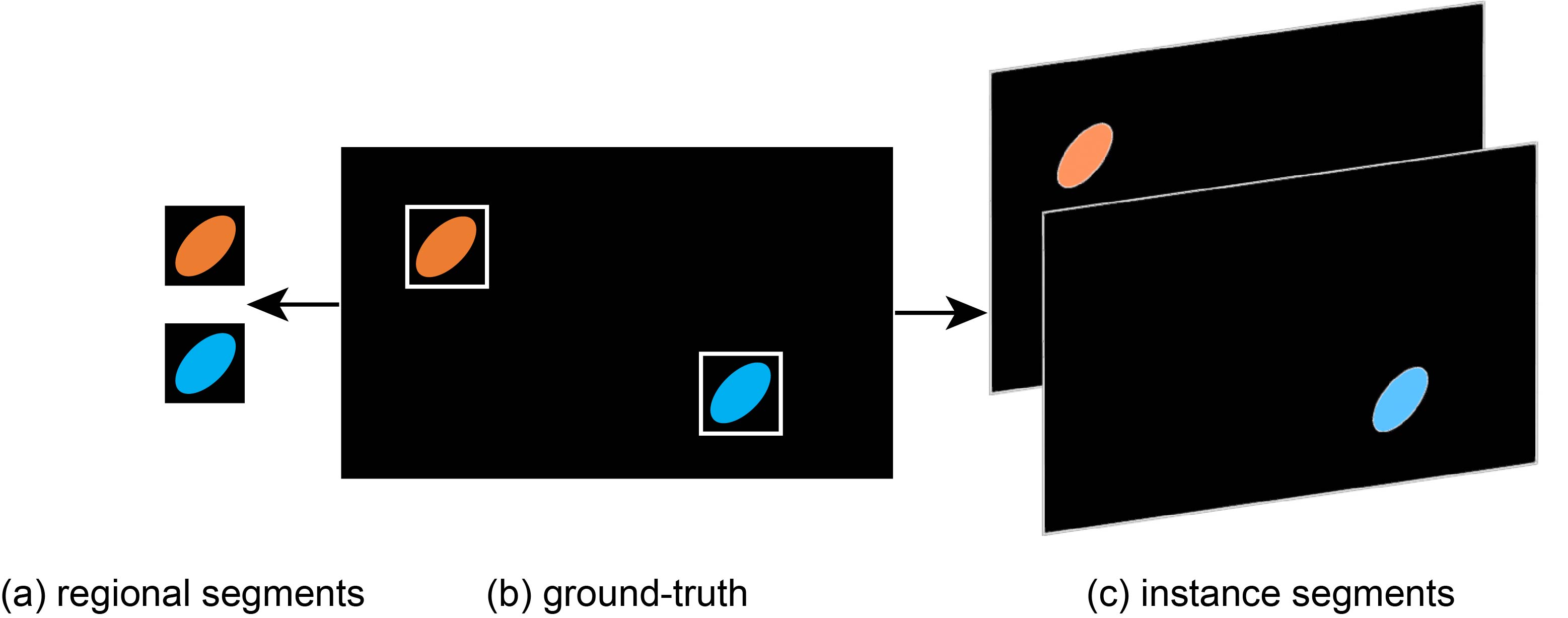}
	\caption{Comparison between regional segments and instance segments.  (a) cropped instance maps from bounding boxes tightly surrounding each object; (b) ground-truth bounding boxes and instance segments within one image; (c) separated entire instance map per object. }
	\label{fig:information_density}
\end{figure}

Recently, transformers have been the prevalent anchor-free approach to address multi-class objects at various scales and shapes. Transformer-based instance segmentation utilizes attention mechanisms and learns latent representative embeddings (queries) from global contextual features to process vastly varying objects. However, existing models \citep{carion2020end, cheng2022masked, shickel2023spatially} are resource-intensive for dense structures in large-scale datasets. That is due to two shortages: (1) a large number of {static} queries from one embedding per object; (2) low instance map occupancy from one object per instance map (depicted in Figure \ref{fig:information_density}c). Hence, they are limited to processing classes with sparse objects, like glomeruli. \par
In response to these challenges, we propose a generic and extensible system for dense instance segmentation and lesion classification on large-scale datasets with potential changes in lesion combinations. Our design has two key components: (a) A novel dense instance segmentation sub-network that recognizes basic anatomical structures, i.e., glomeruli, tubuli, and arteries; (b) a lesion classification sub-network with a set of independent heads that predicts lesions for each class. It is crucial to separately modularize lesion classification and dense instance segmentation for the big picture. The segmentation of basic anatomical structures is a universal foundation for all visual assessment systems. However, lesion classification is a task-driven downstream application, which might vary depending on the interest of users. Therefore, the robustness of dense instance segmentation from local changes in heads of lesion classification is beneficial for the extensibility of our system and for expanding datasets in clinical scenarios. \par  
More specifically, for dense instance segmentation, we propose a novel end-to-end approach, named \textbf{DiffRegFormer}, which effectively combines the advantages of diffusion models, RCNNs, and transformers to tackle dense objects with multi-class and multi-scale. Diffusion models \citep{ho2020denoising} generate bounding box proposals from Gaussian noise, and therefore eliminating the need for pre-defined anchors; RCNNs efficiently crop dense instance maps into dense regions which have very high occupancy rate w.r.t. the bounding boxes ({see} Figure \ref{fig:information_density}. (a)); Cross-attention mechanisms with dynamic queries \citep{chen2021regionvit, li2022dn, zhang2022dino, cheng2022sparse, li2023mask} extract long-range contextual features and enables robust representation of objects across varying scales. DiffRegFormer is not just a simple assembly of the aforementioned techniques. Such assemblies may suffice for sparse instance segmentation on datasets like MSCOCO \citep{lin2014microsoft}. They, however, fail in the context of dense instance segmentation of kidney biopsy ROIs. The crux roots in the proposals that are essentially Gaussian noise at the early training stage of diffusion models. Those noisy candidates, consequently, hinder training due to accumulating errors. Our ablation study analyzes these effects in detail ({see} section \ref{subsubsec:ablation_train_strategies}) and highlights the importance of our specific designs. Specifically, we introduce the following \textbf{key innovations} to address the challenges for dense instance segmentation:
\begin{itemize}
    \item \textbf{Regional features}: Similar to RCNNs, feature maps are cropped using generated proposals and converted to dynamic queries for efficient long-range dependency modeling;  this is robust to large-scale variations. \par
    \item \textbf{Feature disentanglement}: Instead of using shared feature maps, we redesign the model structure to generate separate feature maps for the bounding box decoder and mask decoder; this is crucial to stabilize the training of the mask decoder. \par
    \item \textbf{Class-wise balanced sampling}: Unlike conventional sampling methods, we propose a novel sampling approach. The key difference is to select \textbf{class-wise balanced} positive samples among the \textbf{ground-truth} boxes instead of the proposal ones. \par
\end{itemize}\par
\begin{figure*}[!htbp]
	\centering
	\includegraphics[width=0.95\textwidth, keepaspectratio=true]{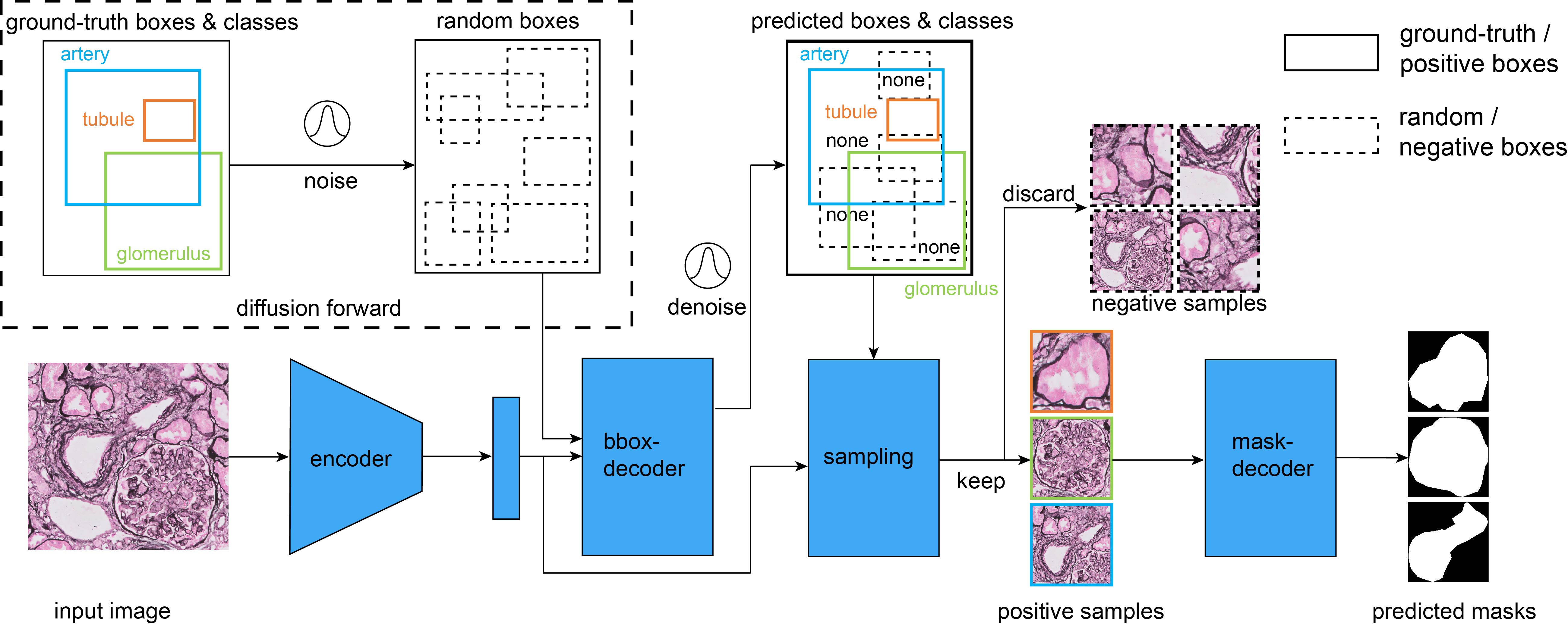}
	\caption{Our \textbf{DiffRegFormer} is a one-stage anchor-free method. Instead of pre-defined anchors, we impose Gaussian noise on ground-truth boxes and generate a fixed-sized set of random bounding boxes (bbox). With feature maps extracted from the encoder, the bbox-decoder iteratively learns to denoise and predicts class-wise candidate boxes. Due to many unevenly distributed candidates, we propose a sampling module that effectively discards negative samples while maintaining a proportion of balanced positive samples for fast convergence. Finally, we make final instance masks according to the selected positive samples in the mask-decoder. For simplicity in illustration, we only choose one anatomical object per class (artery, tubule, glomerulus).}
	\label{fig:flow_chart}
\end{figure*}

In conclusion, based on those key designs, the proposed model is a generic and extensible approach for large-scale datasets with little overhead. To the best of our knowledge, DiffRegFormer is the first end-to-end framework that combines diffusion methodology with a transformer in RCNN-style to process dense objects within ROIs efficiently for multi-scale and multi-class. Figure \ref{fig:flow_chart} depicts a flow chart of the dense instance segmentation sub-network. \par
Each anatomical structure has a dedicated classification head for the lesion classifier since class-wise lesions can differ depending on the dataset's lesion combinations. In Figure \ref{fig:lesion_classifier}, we input the croppings of anatomical structures and resized them to the same size. Each head is trained only with patches of the specific structures and learns to predict multi-label lesions over each class. Our dataset currently only contains sclerotic glomeruli and atrophic tubuli. However, the lesion classifier is extensible to the expanded large-scale dataset at minimal cost because adding, removing, or replacing heads cannot disrupt the other modules. \par
Our work aims to propose a \textbf{generic technical solution} that has flexibility and scalability for automating lesion classification on dense anatomical structures in clinical scenarios. It works as the foundation for further quantification analyses. The main contributions of this work are the following:
\begin{itemize}
    \item {We propose the first end-to-end dense instance segmentation that effectively combines diffusion models, RCNNs, and transformers for multi-class, multi-scale objects. More importantly, our model can directly process ROIs of kidney biopsies, avoiding patch splitting.} 
    \item We propose novel designs to address the accumulative errors caused by diffusion models at an early stage and stabilize the training process. 
    \item We propose a novel class-wise balance sampling method to improve detection and segmentation performance. 
    \item {Instead of the {static} queries used in transformers, we convert regional features into dynamic queries and model the long-range dependencies between regional features while avoiding performance deterioration on dense objects.}
    \item We compare DiffRegFormer with previous models that can process multi-class and multi-scale objects within ROIs in an end-to-end manner. Our model outperforms the previously published models in evaluating Jones' silver-stained images.
    \item We show that DiffRegFormer has the potential of stain (domain)-agnostic detection for PAS-stained images without stain-specific fine-tuning.
    \item {Our lesion classifier can achieve multi-lesion classification. In addition, our plug-and-play strategy can flexibly adapt to large-scale datasets with potential changes in lesion combinations at minimal overhead.}
\end{itemize}\par
The remainder of the paper is organized as follows. In section \ref{sec:relate_work}, all relevant work is introduced, and we elaborate on all improvements w.r.t. previous research. In section \ref{sec:methods}, we describe each component of our model in detail. In section \ref{sec:results}, we demonstrate the extensive evaluation of our model, including comparison experiments and ablation studies. In section \ref{sec:conclusion}, we describe the advantages and limitations of our pipeline and show possible future follow-up research.

\section{Related Work}\label{sec:relate_work}
\noindent \textbf{Semantic Segmentation with Post-Processing}: In the context of lesion classification of dense structures for renal biopsies, semantic segmentation with post-processing has drawn substantial attention. Previous studies \citep{hermsen2019deep, bouteldja2021deep, salvi2021automated, deng2023omni} treat anatomy-wise lesions (e.g., atrophic tubuli) as semantic classes alongside basic anatomical structures (e.g., arteries, glomeruli). An auxiliary \textbf{border} class is often integrated during the training phase to facilitate splitting semantic masks into distinct instances. This approach has demonstrated notable performance. However, challenges are faced in scalability and adaptability to large-scale datasets with potential changes in lesion composition. These issues are at hand as (1) Reliance on ad-hoc procedures (e.g., thresholding and morphological operations \citep{digital_image_processing}) for border generation reduces generalizability. (2) Due to the inflexibility of dataset change, any modification to lesion types necessitates model redesign and retraining. (3) Computational cost scales with lesion classes in the dataset. (4) The difficulty in modeling long-range correlations between objects of varying size due to splitting ROIs into tiles. {(5) It cannot tackle multi-label lesion classification.} In contrast to these issues, our end-to-end framework overcomes these limitations by directly modeling dense objects within ROIs through attention mechanisms. \\\\
\noindent \textbf{Diffusion Model}: Diffusion models, a class of deep generative models  \citep{ho2020denoising, nichol2021improved}, learn to approximate complex distributions through iterative denoising. Despite success in {data generation} {\citep{huang2023stochastic, oh2023diffmix, li2024pathup}}, natural language processing \citep{li2022diffusion}, audio processing \citep{popov2021gradtts}, {and self/weakly-supervised learning \citep{fan2024dcdiff}}, their application to dense instance segmentation remains limited. Initial attempts have been to introduce diffusion models to segmentation tasks \citep{chen2022diffusiondet, gu2022diffusioninst}. The adaptation to dense instance segmentation, however, remains challenging. From our ablation studies, we formulate two primary challenges: (1) The potential dominance of the diffusion modules over feature learning in shared feature maps destabilizes the training of instance prediction modules. (2) Accumulative errors from the diffusion module's early training stages impair the accuracy of dense object predictions. Our approach introduces innovative strategies to overcome these challenges, enabling the first successful implementation of a diffusion model for dense instance segmentation. \\\\
\noindent \textbf{Transformer}: Transformer-based approaches introduce an anchor-free paradigm in an end-to-end manner for instance segmentation, leveraging their capability to capture long-range dependencies and aggregate contextual information. In general, the dependency extraction process is called the attention mechanism, while each context feature is denoted as a query for object representation. Early transformer-based methods \citep{carion2020end, cheng2022masked, shickel2023spatially} rely on {static} queries to model all objects within a dataset. These approaches are incompatible with dense object scenarios due to the inevitable expansion of the {static} query set with dataset size. Recent works show  \citep{li2022dn, zhang2022dino, cheng2022sparse, li2023mask} advancement towards dynamic queries, allowing for temporary modeling of objects within a mini-batch. In this manner, the complexity of processing dense objects is reduced. However, their instance segment representation remains inefficient for dense instance segmentation\footnote{We conducted experiments on our dataset from the official code of both. Compared to our approach, they required more than twice the GPU memory with a batch of size 2 with 500 queries. That is not infeasible for applications with dense objects.} due to the expensive representation of instance segments (Figure \ref{fig:information_density}.(c)) with low occupation rate. Our DiffRegFormer addresses this by integrating region proposals with transformers for improved representation efficiency. \\\\
\noindent \textbf{RCNN}: The Regional Convolutional Neural Network (RCNN) framework \citep{girshick2015fast, ren2016faster} effectively generates candidate regions (i.e. proposals) for dense objects. For instance segmentation, RCNN-based methods used mask modules to predict binary instance masks within proposals. Early variants \citep{he2017mask, huang2019mask, jha2021instance, jiang2021deep} relied on class-agnostic proposals followed by class-specific mask prediction, constrained by the limitations of pre-defined anchors. More recent approaches \citep{fang2021instances} have explored the mapping from proposals to {static} queries, employing cross-attention mechanisms for instance mask prediction. However, this paradigm faces scalability issues in dense object segmentation due to the computational overhead associated with {static} queries ({see} QueryInst in Table \ref{eval_tab_overall}). Our method adopts dynamic queries, balancing computational efficiency and scalability for dense instance segmentation. \par
In summary, our work combines diffusion modeling and attention mechanisms into an RCNN-style framework, which addresses challenges of scalability, adaptability, and efficiency inherent in existing approaches for dense instance segmentation in renal biopsies.
\section{Methods}\label{sec:methods}
\subsection{Preliminaries}\label{subsec:preliminaries}
\noindent \textbf{Diffusion model}. Diffusion approaches \citep{song2020denoising, rombach2022high} is a family of deep generative models inspired by the principles of non-equilibrium thermodynamics \citep{sohl2015deep, huang2023stochastic}. Their operation is conceptualized as $T$-step sequential process, iteratively transiting from an initial state $Z_{0}$ to a final state of pure noise $Z_{T}$. Diffusion forward involves the progressive addition of Gaussian noise at each transition. Due to the special properties of the Gaussian distribution \citep{ho2020denoising}, it is possible to sample any intermediate noisy at state $Z_t$ directly:
\begin{equation}\label{eq:diffusion_forward}
	q(Z_{t}|Z_{0}) = \mathcal{N}(Z_{t} | \sqrt{\bar{\alpha}_{t}}Z_{0}, (1-\bar{\alpha}_{t} \mathbf{I}))
\end{equation}
where $Z_{0}$ is the original data, $Z_{t}\ (t \le T)$ is the latent state at step $t$, and $\bar{\alpha}_{t}$ is the cumulative noise variance schedule \citep{chen2022diffusiondet}. Notably, the diffusion forward process does not introduce trainable parameters. The objective of a diffusion model is to learn the reverse process. A neural network, $f_{\theta}(Z_{t}, t)$, approximates the recovery of the original distribution $Z_{0}$ from a noisy state $Z_{t}$. Training optimizes this network by minimizing a loss function to denoise at arbitrary intermediate state $t$:
\begin{equation}
	\mathcal{L} = \frac{1}{2}\|f_{\theta}(Z_{t}, t) - Z_{0}\|^{2}
\end{equation}
where $Z_{0}$ represents the ground-truth and $Z_{t}$ is the noisy input. During inference, the model $f_{\theta}$ reconstructs from pure noise $Z_{T}$ to the predicted data $Z_{0}$ through a series of reverse updates \citep{song2020denoising} with a step-size $s$: $Z_{T} \rightarrow Z_{T-s} \rightarrow \cdots \rightarrow Z_{0}$. See \citep{song2020denoising} for a detailed derivation.
\subsection{Dense Instance Segmentation Model}\label{subsec:instance_segmentation_model}
The iterative proposal generation in DiffRegFormer necessitates multiple model executions at inference. That poses a computational burden if the entire model runs on the raw image at each step during training. To address this issue, we decouple the image encoder from the remaining modules. This strategy enables the encoder to extract multi-scale feature maps from the raw input image only once. Then, subsequent modules operate for iterative refinement of box and mask predictions from an initial state of Gaussian noise conditioned on these deep features. Moreover, to facilitate the mask-decoder training, we introduce a novel sampling module that randomly selects positive samples from each class, ensuring a balanced distribution of class-wise regional features. That facilitates learning object representations with lower frequencies in the dataset. \par
In summary, the DiffRegFormer comprises four components:
\begin{itemize}
    \item \textbf{Encoder}: Processes the raw input image once to extract multi-scale feature maps for bbox-decoder and mask-decoder separately.
    \item \textbf{Bbox-decoder}: Utilizes the extracted feature maps to iteratively refine bounding box predictions.
    \item \textbf{Sampling Module}: Stabilizes the training by randomly selecting class-wise balanced positive samples for the mask-decoder.
    \item \textbf{Mask-decoder}: Iteratively refines mask predictions based on the sampled regional features.
\end{itemize}
\subsubsection{Encoder}
The image encoder extracts high-level feature maps from the raw image for subsequent decoders. To generate multi-scale feature maps, it utilizes a ResNet \citep{he2015deep} pre-trained on ImageNet \citep{deng2009imagenet} as its backbone, followed by a Feature Pyramid Network (FPN) \citep{lin2017feature}. After FPN, a stack of $3\times 3$ convolutional operations is applied to produce two separate feature maps, thereby decoupling the bbox-decoder and mask-decoder. \par
\subsubsection{Bbox-decoder}
\begin{figure}[!htbp]
	\centering
	\includegraphics[width=0.5\textwidth, keepaspectratio=true]{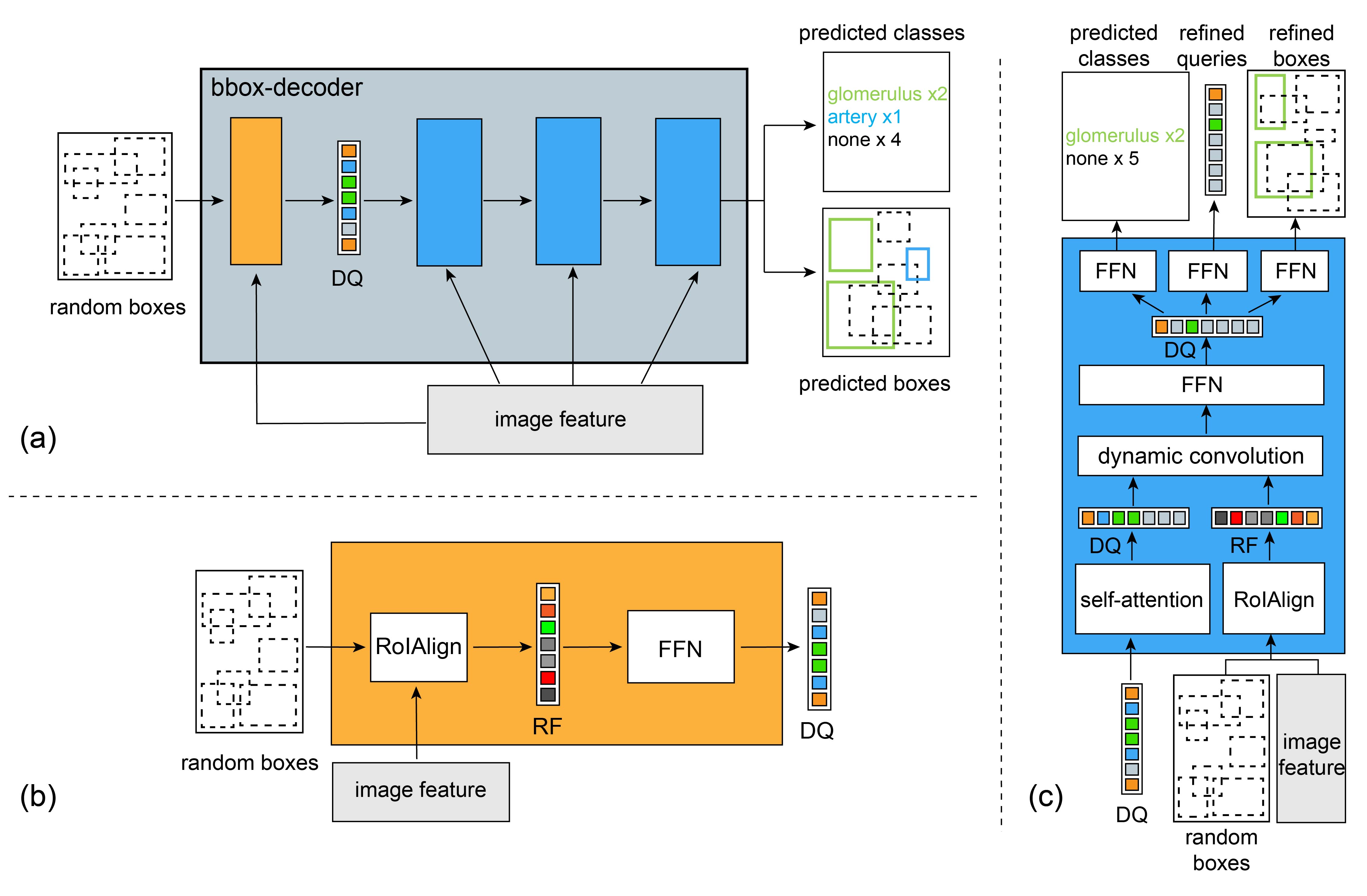}
	\caption{The bounding box (bbox)-decoder takes multi-scale feature maps and a set of random boxes as input. Then, the prediction of classes and boxes will be outputted iteratively. (a) The module comprises an initialization head for dynamic queries (orange rectangle) and multiple box refinement heads (blue rectangles). (b) In the orange rectangle in (a), the initial dynamic queries are generated via a RoIAlign pooling operator and a feed-forward network (FFN). (c) Each box refinement module (one blue rectangle in (a)) takes the previous stage's dynamic queries and proposal boxes as input, generating predictions and refined dynamic queries for the next stage. Abbreviations: dynamic queries: DQ, feed-forward network: FFN, regional features: RF.}
	\label{fig:box_decoder}
\end{figure}

The bbox-decoder builds upon DiffusionDet \citep{chen2022diffusiondet}. As shown in Figure \ref{fig:box_decoder}.(a), a set of random boxes and multi-scaled feature maps is first taken from the encoder as input. Then, dynamic queries (Figure \ref{fig:box_decoder}b) are initialized from a regional pooling operator (RoIAlign \citep{he2017mask}) and a feed-forward network (FFN) \citep{vaswani2017attention}. These queries undergo iterative refinement through multiple stages. Each stage takes proposal boxes and dynamic queries from the previous stage, generating refined dynamic queries, box predictions, and class predictions for the next stage. Figure \ref{fig:box_decoder}c illustrates one detailed refinement stage. First, new regional features (cropped by RoIAlign) interact with dynamic queries (after a self-attention module) via dynamic convolution \citep{chen2020dynamic}, emphasizing regions likely containing objects while suppressing others. Next, the enhanced regional features are processed by FFNs to produce refined box predictions, box-wise classifications, and queries, respectively. \par
\subsubsection{Sampling}
Conventional RCNN models use intersection over union (IoU) to classify a proposal box as positive if its IoU with a ground-truth box is $\geq 0.5$ \citep{he2017mask}. This approach, however, is ill-suited for diffusion methods, particularly in early training stages when diffusion methods predominantly produce noise. That can complicate convergence or even cause failure due to error accumulation. Further, conventional IoU-based methods are susceptible to class imbalances, biasing the learning process towards frequently occurring objects, thereby neglecting rare instances. We, therefore,  propose a novel sampling method to address these issues. We first replace proposal boxes with ground-truth boxes, ensuring a sufficient supply of positive samples. That is justified because the bbox-decoder aims to predict boxes that closely approximate the ground-truth. We then divide ground-truth bounding boxes into groups based on their classes. Within each group, we randomly select up to $n$ boxes as positive samples without repetition. {The core idea is that objects of each class having similar shapes and appearances can be mapped to a cluster. Consequently, it allows us to learn instance feature representations with a small sample number $n$. This especially applies to the tubular class that occupies most of our dataset.} This sampling strategy offers three distinct advantages:
\begin{enumerate}
    \item \textbf{Rare Instance Learning}: The mask-decoder can effectively learn representations of infrequently occurring instances whenever they appear in the input, mitigating the risk of being overwhelmed by more common classes.
    \item \textbf{Randomize Frequent Instance}: Learning on a randomized subset of frequent instances prevents model over-fitting. 
    \item \textbf{Training Stability}: Employing selected ground-truth boxes as positive samples promotes stability and accelerates mask training.
\end{enumerate}
\subsubsection{Mask decoder}
\begin{figure}[!htbp]
	\centering
	\includegraphics[width=0.5\textwidth, keepaspectratio=true]{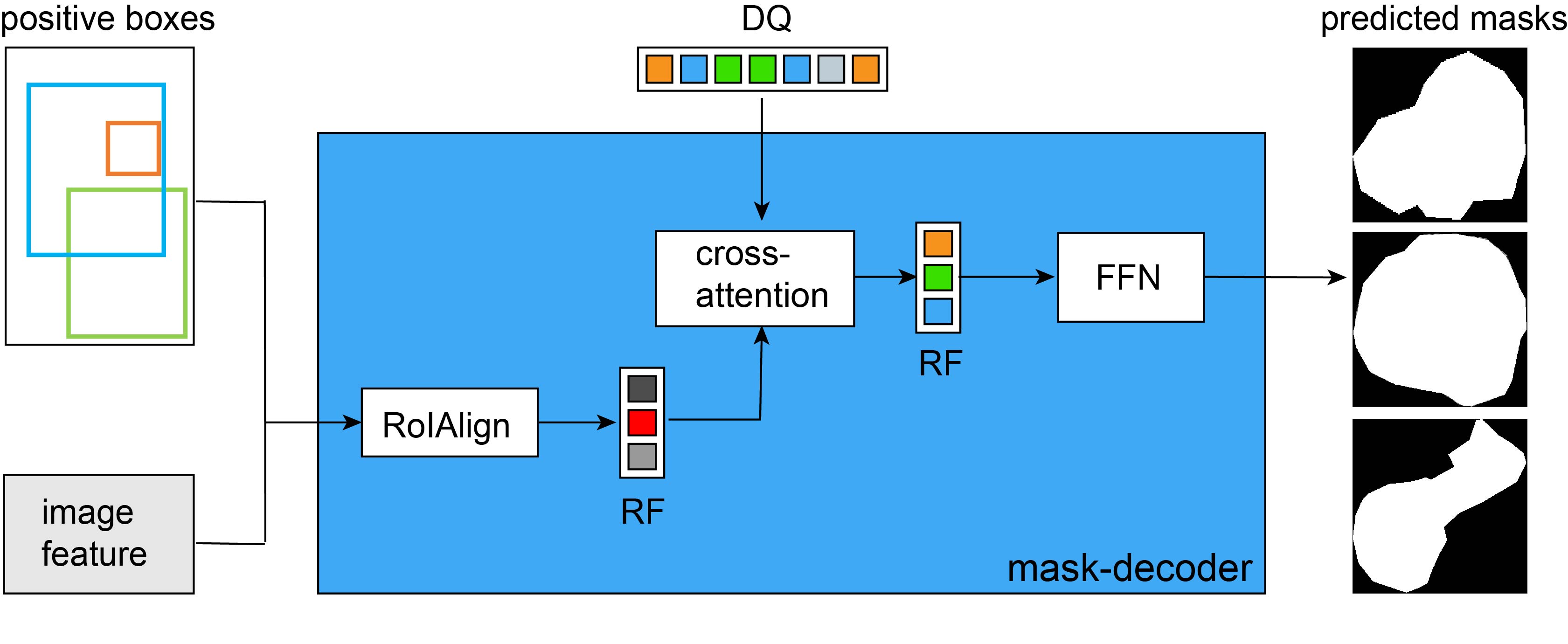}
	\caption{The mask-decoder takes multi-scale feature maps and a set of positive boxes as input and predicts instance masks. The dynamic queries interact with regional features using cross-attention and only highlight pixels residing in proposal boxes. Final instance masks are generated from the enhanced regional feature maps. Abbreviations: dynamic queries: DQ, feed-forward network: FFN, regional features: RF.}\label{fig:mask_decoder}
\end{figure}

The mask decoder iteratively refines mask predictions, utilizing regional features cropped from multi-scale feature maps based on boxes selected by the sampling module. Instead of using {static} queries \citep{cheng2022masked,cheng2021per}, we compute cross-attention between dynamic queries and positive regional features. That highlights effective queries and filters others, enhancing regional features by focusing on pixels within positive regions, thereby facilitating accurate per-pixel instance score prediction (Figure \ref{fig:mask_decoder}). \par
\begin{figure}[!htbp]
	\centering
	\includegraphics[width=0.5\textwidth, keepaspectratio=true]{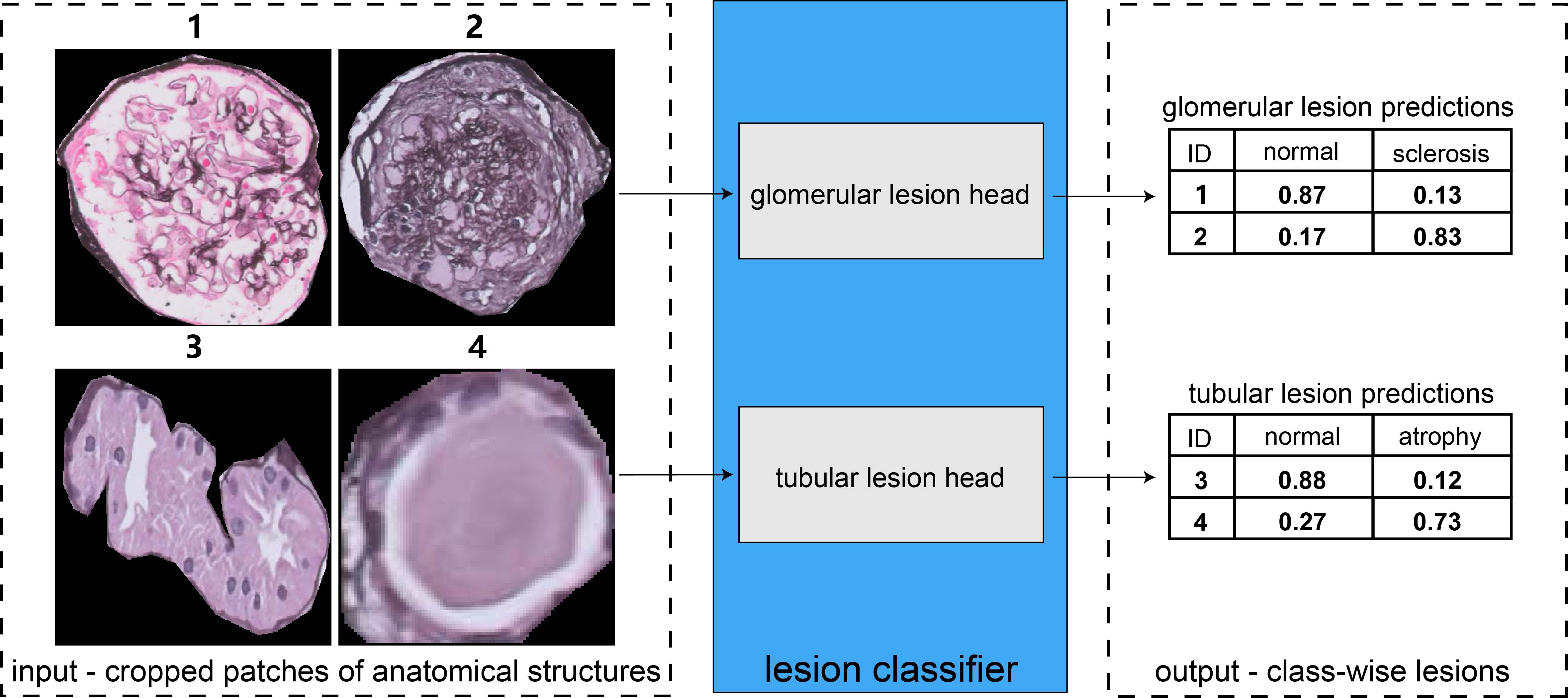}
	\caption{The lesion classifier consists of 2 prediction heads, one dedicated to glomeruli and the other to tubuli. It takes croppings of anatomical structures and outputs the probability of class-wise lesions. }
	\label{fig:lesion_classifier}
\end{figure}
\begin{figure}[!htbp]
	\centering
	\includegraphics[width=0.5\textwidth, keepaspectratio=true]{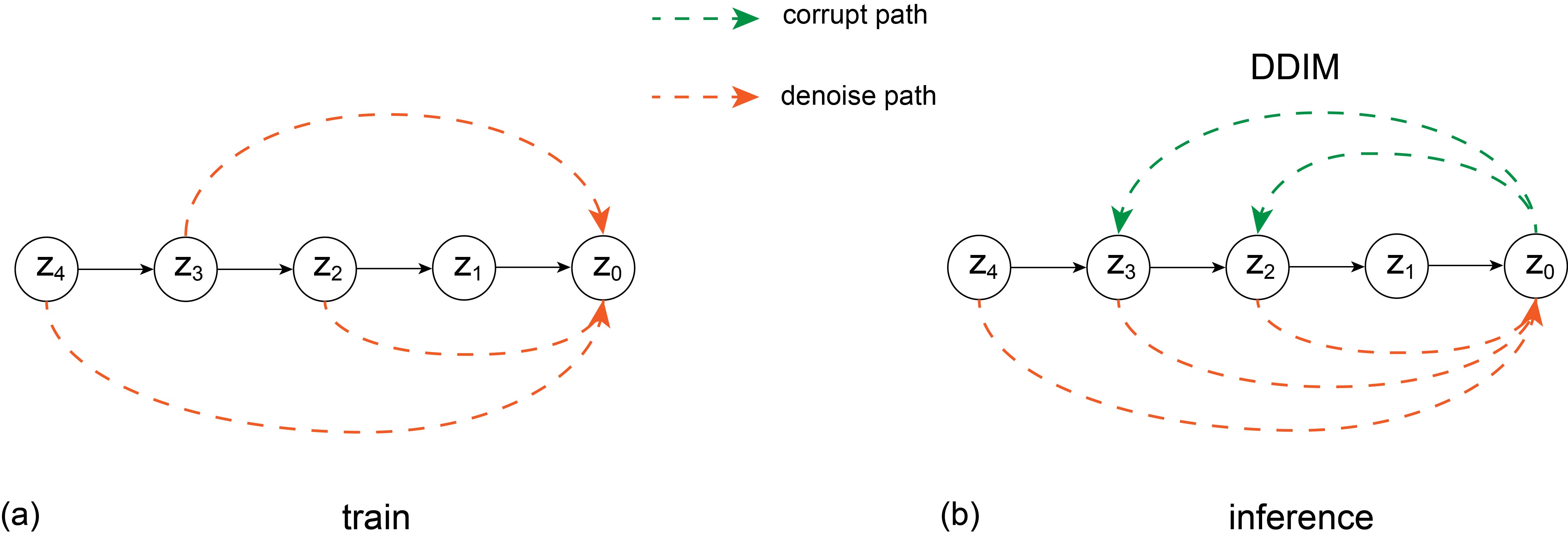}
	\caption{An illustration for a diffusion model at different stages depicts a simplified process with $5$ states (skip step=$1$), where state $4$ is pure Gaussian noise and state $0$ is the ground-truth. (a) During training, a diffusion model randomly selects denoising paths to train its decoder; (b) In inference, the Denoising Diffusion Implicit Model (DDIM) flexibly alternates between denoising and corrupting paths, generating multiple predictions that are then ensembled for a refined final result.}
	\label{fig:duffusion_process}
\end{figure}
\subsection{Lesion classifier}\label{subsec:lesion_classifier}
The lesion classifier comprises multiple independent lesion prediction heads. Each head specializes in a specific anatomical structure (e.g., glomerulus) and processes cropped images of the specified structure to predict probabilities for multi-label lesions. Figure \ref{fig:lesion_classifier} illustrates this design. The architecture of a single head consists of three convolutional blocks followed by a max-pooling operator. Each block contains a $3 \times 3$ convolutional, batch normalization, and ReLU activation. We currently implement two prediction heads for glomeruli (predicting sclerosis) and tubuli (predicting atrophy). \par
Notably, the modular design of the instance segmentation component allows lesion prediction heads to operate as independent plug-ins. This modularity is crucial for seamlessly adapting the model to the expanding datasets, particularly in clinical scenarios. It allows for the straightforward integration or removal of prediction heads with minimal impact on the overall model architecture. For example, if new annotations for arteries become available, a dedicated prediction head can be trained to classify arterial lesions and integrate them into the model without modifying existing components. Likewise, this design extends the model's diagnostic scope to include additional lesions within a given structure (e.g., tubuli) by simply replacing an existing prediction head with a newly trained one on expanded data. This design promotes the model's versatility and long-term adaptability within clinical scenarios where the scope of identifiable lesions may increase.
\subsection{Implementation Details}\label{subsec:implementation}
The entire model adopts distinct strategies for the training and inference phases. During training, the DiffRegFormer learns to reconstruct ground-truth boxes through adaptation at state $0$ from an arbitrary noisy state $t$ (Figure \ref{fig:duffusion_process}.(a)). Additionally, the sampling module provides class-balanced ground-truth boxes to facilitate the mask decoder's training. Lastly, The lesion classifier operates on croppings generated from ground-truth instance masks. Each prediction head specializes in classifying lesions specific to the corresponding anatomical structure within these croppings. At the same time, during inference, the DiffRegFormer assembles multiple predictions to refine final results (Figure \ref{fig:duffusion_process}b). Moreover, the sampling module is deactivated, and all proposals are directly forwarded to the mask decoder for instance segmentation prediction. Likewise, the lesion classifier utilizes instance masks predicted by DiffRegFormer to crop structures. As a last step, each prediction head predicts probabilities for class-specific lesions within these cropped regions.
\begin{algorithm}[!htbp]
	\caption{DiffRegFormer Training}\label{algorithm:training}
	\begin{lstlisting}[language=python]
def train(images, gt_boxes, gt_masks):
  """
  images: [B, H, W, 3]
  gt_boxes: [B, *, 4]
  gt_masks: [B, H, W, G]
  B: batch size
  N: number of proposal boxes
  """
  # generate multi-scale features via encoder
  feats = encoder(images)
  # generate noised boxes at state t where t 
  # is a random integer in diffusion forward; 
  # Pad boxes to N with t_boxes: [B, N, 4]
  t_boxes, t = diffusion(gt_boxes, mean=0, std=1)
  # generate initial dynamic queries
  d_query = intialize_query(t_boxes, feats)
  # learn to reverse noised boxes at state t back
  # to ground-truth boxes at state 0 and return
  # refined dynamic queries
  [pred_boxes, d_query] = box_decoder(t_boxes, feats, d_query, t)
  # obtain bbox loss via the set objective function
  loss_bbox = set_loss(pred_boxes, gt_boxes)
  # randomly select balanced positive boxes from
  # groud-truth boxes
  pos_boxes = sampling(gt_boxes)
  # generate predicted instance masks
  pred_masks = mask_decoder(pos_boxes, feats, d_query)
  # obtain mask loss via the set objective function
  loss_mask = set_prediction_loss(pred_masks, gt_masks)
  return loss_bbox, loss_mask
	\end{lstlisting}
\end{algorithm}
\subsubsection{Training}
We start by constructing noisy boxes from the ground-truth. The DiffRegFormer is then trained to recover bounding boxes and iteratively generate instance masks from an arbitrary state $t$. The algorithm is detailed in {Algorithm} \ref{algorithm:training}. \\\\
\noindent \textbf{Diffusion forward}. Ground-truth boxes are first concatenated with additional random boxes to align to a fixed length across images, accounting for the variability in instance counts. Gaussian noise is then introduced as the noise scale governed by $\bar\alpha$ as per Eq.\eqref{eq:diffusion_forward}. Following \citep{nichol2021improved}, the scale varies with state $t$ using a monotonically decreasing \textit{cosine} schedule. It should be noted that, for optimal performance, the signal-to-noise ratio requires a relatively high signal scaling value compared to image generation tasks \citep{chen2022generalist, ho2020denoising, dhariwal2021diffusion, chen2022analog, chen2022diffusiondet}. \\\\
\noindent \textbf{Training Loss}. The box decoder inputs noisy boxes and predicts category classifications and box coordinates. Given the one-to-many relationship between ground-truth and predicted boxes, first, a set-loss function is employed to match predictions to the ground-truth based on the lowest cost \citep{carion2020end,sun2021makes,zhu2020deformable}. Furthermore, the mask decoder inputs sampled positive boxes and predicts instance masks. We use a Binary Cross Entropy loss function due to the one-to-one mapping. Finally, since multiple lesions can exist within one cropping, lesion classification adopts a Binary Cross Entropy loss function for multi-label prediction within each head. \\\\
\noindent \textbf{Sampling}. Our implementation uses $3$ classes and a maximum of $N=500$ objects per image. We set up to $n=\frac{500}{3}\simeq 166$ instances per class. More specifically, positive ground-truth boxes are sampled without repetition, not exceeding the class-wise maximum. Moreover, we disregard absent classes within the input. For example, if there are 2 glomeruli, 0 vessels, and 200 tubuli ground-truth boxes within an ROI, then the sampling module would select 2 glomeruli, 0 vessels, and 166 random tubuli as positive samples without duplication.
\begin{algorithm}[!htbp]
	\caption{DiffRegFormer Inference}\label{algorithm:inference}
	\begin{lstlisting}[language=python]
def infer(images, step, T):
  """
  images: [B, H, W, 3]
  step: the skip length for the state transform
  T: the length of the chain
  B: batch size
  N: number of proposal boxes
  """
  # generate multi-scale features via encoder
  feats = encoder(images)
  # return Gaussian noise as noisy boxes at state T;  
  # Pad boxes to N with T_boxes: [B, N, 4]
  [t_boxes, _] = diffusion(mean=0, std=1)
  # generate state transform pairs skipping every step 
  # [(T, T-step), (T-step, T-2*step), ..., (step, 0)]
  time_pairs = uniform(0, T, step)
  # generate initial dynamic queries
  d_query = intialize_query(t_boxes, feats)
  # iterate over stages
  for (t_now, t_next) iterate t_paris:
    # predict boxes and dynamic query at state t_now
    [pred_boxes, d_query] = box_decoder(t_boxes, feats, d_query, t_now)
    # generate new noisy boxes from state t_now to t_next
    t_boxes = ddim(t_boxes, pred_boxes, t_now, t_next)
    # replace undesired boxes with random Gaussian noise
    t_boxes = box_repalce(t_boxes)
  # generate predicted instance masks
  pred_masks = mask_decoder(pred_boxes, feats, d_query)
  return pred_boxes, pred_masks
	\end{lstlisting}
\end{algorithm}
\subsubsection{Inference}
During inference, the DiffRegFormer starts from a standard Gaussian distribution corresponding to the final state $T$ as defined in Eq.(\ref{eq:diffusion_forward}). Then, a progressive denoising operation reverses the predictions to the initial state $0$. The algorithm, as depicted in {Algorithm} \ref{algorithm:inference}, outlines this process. \\\\
\noindent \textbf{Inference steps}. Each inference iteration involves two primary operations. First, the box decoder processes noisy boxes generated from the previous state $\mathrm{t_{now}}$ and predicts both categories and box coordinates. Subsequently, a Denoising Diffusion Implicit Model (DDIM) \citep{mohamed2016learning,song2020denoising} introduces noise into the previously predicted boxes. This step generates new noisy boxes for subsequent state $\mathrm{t_{next}}$, facilitating a progressive denoising process across different states. Notably, each iteration performs a single denoising operation, transitioning from state $\mathrm{t_{now}}$ to $\mathrm{t_{next}}$. This characteristic allows for the assembly of intermediate predictions from multiple denoising steps (i.e., $t_{i}\to 0$ where $i\in [1, T]$) to refine the outcome. However, DiffRegFormer performs inference only once to improve efficiency, directly transitioning from pure Gaussian noise at state $T$ to state $0$ to obtain the final predictions. \\\\
\noindent \textbf{Box replacement}. Predicted boxes can be categorized as either \textbf{\textit{positive}} (scores above a threshold, containing objects of interest) or \textbf{\textit{negative}} (scores below the threshold, arbitrarily located). Directly applying negative boxes into DDIM would significantly degrade the quality of newly generated noisy boxes. This issue stems from the fact that negative boxes, originating from corrupted boxes during the training phase, significantly deviate from the Gaussian distribution. To ensure consistency between the training and inference phases, we substitute negative boxes with random boxes sampled from a Gaussian distribution. \par
\begin{table*}[!ht]
	\caption{Overall evaluation results\label{eval_tab_overall}. {All models use ResNet-50 as a backbone network and have been trained 40000 iterations on an NVIDIA GeForce RTX 3090 GPU with a batch size of 2. AP$_{\mathrm{S}}$, AP$_{\mathrm{M}}$, and AP$_{\mathrm{L}}$ correspond to AP on small, medium, and large-sized objects, respectively. The best results are highlighted in bold. CMask-RCNN represents Cascade Mask-RCNN.}}
    \begin{minipage}{\textwidth}
        \centering
        \begin{tabular}{|p{3mm}<{\centering}p{22mm}||p{4.5mm}<{\centering}p{4.mm}<{\centering}p{4.5mm}<{\centering}p{4.5mm}<{\centering}p{4.5mm}<{\centering}p{4.5mm}<{\centering}|p{4.5mm}<{\centering}p{4.5mm}<{\centering}p{4.5mm}<{\centering}p{4.5mm}<{\centering}p{4.5mm}<{\centering}p{4.5mm}<{\centering}|}
        \hline
        \multicolumn{2}{|c||}{\multirow{2}{*}{}} & \multicolumn{6}{c|}{Bounding Boxes} & \multicolumn{6}{c|}{Instances} \\
        \cline{3-14}\multicolumn{2}{|c||}{} & \multicolumn{1}{c|}{AP} & \multicolumn{1}{c|}{AP$_{50}$} & \multicolumn{1}{c|}{AP$_{75}$} & \multicolumn{1}{c|}{AP$_{\mathrm{S}}$}  & \multicolumn{1}{c|}{AP$_{\mathrm{M}}$}  & AP$_{\mathrm{L}}$  & \multicolumn{1}{c|}{AP}   & \multicolumn{1}{c|}{AP$_{50}$} & \multicolumn{1}{c|}{AP$_{75}$} & \multicolumn{1}{c|}{AP$_{\mathrm{S}}$}  & \multicolumn{1}{c|}{AP$_{\mathrm{M}}$}  & AP$_{\mathrm{L}}$  \\ \hline
        \multicolumn{1}{|c|}{\multirow{2}{*}{one-stage}} & QueryInst\_300 & \multicolumn{1}{c|}{37.8} & \multicolumn{1}{c|}{59.3} & \multicolumn{1}{c|}{47.4} & \multicolumn{1}{c|}{13.5} & \multicolumn{1}{c|}{21.5} & 47.4 & \multicolumn{1}{c|}{38.9} & \multicolumn{1}{c|}{59.4} & \multicolumn{1}{c|}{43.5} & \multicolumn{1}{c|}{8.8}  & \multicolumn{1}{c|}{19.0} & 44.1 \\ \cline{2-14} 
        \multicolumn{1}{|c|}{}                           & QueryInst\_500       & \multicolumn{1}{c|}{30.1} & \multicolumn{1}{c|}{31.5} & \multicolumn{1}{c|}{22.8} & \multicolumn{1}{c|}{0.9}  & \multicolumn{1}{c|}{8.7}  & 24.1 & \multicolumn{1}{c|}{32.5} & \multicolumn{1}{c|}{31.9} & \multicolumn{1}{c|}{22.7} & \multicolumn{1}{c|}{0.4}  & \multicolumn{1}{c|}{7.8}  & 27.7 \\ \hline
        \multicolumn{1}{|c|}{\multirow{2}{*}{two-stage}} & Mask-RCNN            & \multicolumn{1}{c|}{49.0} & \multicolumn{1}{c|}{68.6} & \multicolumn{1}{c|}{56.2} & \multicolumn{1}{c|}{\textbf{18.0}} & \multicolumn{1}{c|}{24.4} & 59.2 & \multicolumn{1}{c|}{45.5} & \multicolumn{1}{c|}{69.0} & \multicolumn{1}{c|}{53.5} & \multicolumn{1}{c|}{\textbf{16.4}} & \multicolumn{1}{c|}{21.5} & 54.4 \\ \cline{2-14} 
        \multicolumn{1}{|c|}{}                           & CMask-RCNN    & \multicolumn{1}{c|}{49.9} & \multicolumn{1}{c|}{67.3} & \multicolumn{1}{c|}{\textbf{58.4}} & \multicolumn{1}{c|}{7.6}  & \multicolumn{1}{c|}{24.1} & \textbf{61.4} & \multicolumn{1}{c|}{45.0} & \multicolumn{1}{c|}{67.8} & \multicolumn{1}{c|}{\textbf{54.5}} & \multicolumn{1}{c|}{6.5}  & \multicolumn{1}{c|}{21.6} & \textbf{54.5} \\ \hline
        \multicolumn{1}{|c|}{our model}                  & DiffRegFormer & \multicolumn{1}{c|}{\textbf{52.1}} & \multicolumn{1}{c|}{\textbf{71.1}} & \multicolumn{1}{c|}{57.7} & \multicolumn{1}{c|}{15.4} & \multicolumn{1}{c|}{\textbf{27.4}} & \textbf{61.4} & \multicolumn{1}{c|}{\textbf{46.8}} & \multicolumn{1}{c|}{\textbf{71.6}} & \multicolumn{1}{c|}{52.8} & \multicolumn{1}{c|}{14.1} & \multicolumn{1}{c|}{\textbf{23.6}} & 54.2 \\ \hline
        \end{tabular}
    \end{minipage}
\end{table*}

\begin{table*}[!htbp]
    \centering
	\caption{Per-class evaluation results\label{eval_tab_percat}. {All models use ResNet-50 as a backbone network and have been trained 40000 iterations on one NVIDIA GeForce RTX 3090 GPU with a batch size of 2. AP$_{\mathrm{S}}$, AP$_{\mathrm{M}}$, and AP$_{\mathrm{L}}$ correspond to mAP on small, medium, and large-sized objects, respectively. The best results are highlighted in bold. CMask-RCNN represents Cascade Mask-RCNN.}}
		\begin{tabular}{|p{3mm}<{\centering}p{22mm}||p{4.5mm}<{\centering}p{4.5mm}<{\centering}p{4.5mm}<{\centering}p{4.5mm}<{\centering}p{4.5mm}<{\centering}p{4.5mm}<{\centering}|p{4.5mm}<{\centering}p{4.5mm}<{\centering}p{4.5mm}<{\centering}p{4.5mm}<{\centering}p{4.5mm}<{\centering}p{4.5mm}<{\centering}|}
            \hline
            \multicolumn{2}{|c||}{\multirow{2}{*}{}} & \multicolumn{6}{c|}{Bounding Boxes - Glomeruli} & \multicolumn{6}{c|}{Instances - Glomeruli} \\ 
            \cline{3-14}\multicolumn{2}{|c||}{} & \multicolumn{1}{c|}{AP}   & \multicolumn{1}{c|}{AP$_{50}$} & \multicolumn{1}{c|}{AP$_{75}$} & \multicolumn{1}{c|}{AP$_{\mathrm{S}}$} & \multicolumn{1}{c|}{AP$_{\mathrm{M}}$} & AP$_{\mathrm{L}}$  & \multicolumn{1}{c|}{AP}   & \multicolumn{1}{c|}{AP$_{50}$} & \multicolumn{1}{c|}{AP$_{75}$} & \multicolumn{1}{c|}{AP$_{\mathrm{S}}$} & \multicolumn{1}{c|}{AP$_{\mathrm{M}}$} & AP$_{\mathrm{L}}$  \\ \hline
            \multicolumn{1}{|c|}{\multirow{2}{*}{one-stage}} & QueryInst\_300       & \multicolumn{1}{c|}{62.6} & \multicolumn{1}{c|}{74.5} & \multicolumn{1}{c|}{71.6} & \multicolumn{1}{c|}{-}   & \multicolumn{1}{c|}{-}   & 63.7 & \multicolumn{1}{c|}{56.1} & \multicolumn{1}{c|}{74.5} & \multicolumn{1}{c|}{68.1} & \multicolumn{1}{c|}{-}   & \multicolumn{1}{c|}{-}   & 57.1 \\ \cline{2-14} 
            \multicolumn{1}{|c|}{}                           & QueryInst\_500       & \multicolumn{1}{c|}{36.7} & \multicolumn{1}{c|}{47.5} & \multicolumn{1}{c|}{44.0} & \multicolumn{1}{c|}{-}   & \multicolumn{1}{c|}{-}   & 37.6 & \multicolumn{1}{c|}{35.9} & \multicolumn{1}{c|}{47.6} & \multicolumn{1}{c|}{43.6} & \multicolumn{1}{c|}{-}   & \multicolumn{1}{c|}{-}   & 37.3 \\ \hline
            \multicolumn{1}{|c|}{\multirow{2}{*}{two-stage}} & Mask-RCNN            & \multicolumn{1}{c|}{69.4} & \multicolumn{1}{c|}{85.8} & \multicolumn{1}{c|}{78.6} & \multicolumn{1}{c|}{-}   & \multicolumn{1}{c|}{-}   & 70.4 & \multicolumn{1}{c|}{67.0} & \multicolumn{1}{c|}{85.8} & \multicolumn{1}{c|}{78.8} & \multicolumn{1}{c|}{-}   & \multicolumn{1}{c|}{-}   & 68.1 \\ \cline{2-14} 
            \multicolumn{1}{|c|}{}                           & CMask-RCNN    & \multicolumn{1}{c|}{70.7} & \multicolumn{1}{c|}{84.0} & \multicolumn{1}{c|}{81.7} & \multicolumn{1}{c|}{-}   & \multicolumn{1}{c|}{-}   & 72.1 & \multicolumn{1}{c|}{65.1} & \multicolumn{1}{c|}{84.0} & \multicolumn{1}{c|}{79.8} & \multicolumn{1}{c|}{-}   & \multicolumn{1}{c|}{-}   & 66.4 \\ \hline
            \multicolumn{1}{|c|}{our model}                  & DiffRegFormer & \multicolumn{1}{c|}{\textbf{80.9}} & \multicolumn{1}{c|}{\textbf{94.2}} & \multicolumn{1}{c|}{\textbf{89.8}} & \multicolumn{1}{c|}{-}   & \multicolumn{1}{c|}{-}   & \textbf{80.9} & \multicolumn{1}{c|}{\textbf{74.3}} & \multicolumn{1}{c|}{\textbf{94.2}} & \multicolumn{1}{c|}{\textbf{80.9}} & \multicolumn{1}{c|}{-}   & \multicolumn{1}{c|}{-}   & \textbf{74.3} \\ \hline
        \end{tabular}
		\begin{tabular}{|p{3mm}<{\centering}p{22mm}||p{4.5mm}<{\centering}p{4.5mm}<{\centering}p{4.5mm}<{\centering}p{4.5mm}<{\centering}p{4.5mm}<{\centering}p{4.5mm}<{\centering}|p{4.5mm}<{\centering}p{4.5mm}<{\centering}p{4.5mm}<{\centering}p{4.5mm}<{\centering}p{4.5mm}<{\centering}p{4.5mm}<{\centering}|}
            \hline
            \multicolumn{2}{|c||}{\multirow{2}{*}{}}                                 & \multicolumn{6}{c|}{Bounding Boxes - Arteries}                                                                                                   & \multicolumn{6}{c|}{Instances - Arteries}                                                                                                        \\ \cline{3-14} 
            \multicolumn{2}{|c||}{}                                                  & \multicolumn{1}{c|}{AP}   & \multicolumn{1}{c|}{AP$_{50}$} & \multicolumn{1}{c|}{AP$_{75}$} & \multicolumn{1}{c|}{AP$_{\mathrm{S}}$}  & \multicolumn{1}{c|}{AP$_{\mathrm{M}}$}  & AP$_{\mathrm{L}}$  & \multicolumn{1}{c|}{AP}   & \multicolumn{1}{c|}{AP$_{50}$} & \multicolumn{1}{c|}{AP$_{75}$} & \multicolumn{1}{c|}{AP$_{\mathrm{S}}$}  & \multicolumn{1}{c|}{AP$_{\mathrm{M}}$}  & AP$_{\mathrm{L}}$  \\ \hline
            \multicolumn{1}{|c|}{\multirow{2}{*}{one-stage}} & QueryInst\_300       & \multicolumn{1}{c|}{18.8} & \multicolumn{1}{c|}{37.0} & \multicolumn{1}{c|}{19.1} & \multicolumn{1}{c|}{5.2}  & \multicolumn{1}{c|}{20.2} & 20.6 & \multicolumn{1}{c|}{17.5} & \multicolumn{1}{c|}{36.3} & \multicolumn{1}{c|}{14.8} & \multicolumn{1}{c|}{4.5}  & \multicolumn{1}{c|}{17.9} & 19.4 \\ \cline{2-14} 
            \multicolumn{1}{|c|}{}                           & QueryInst\_500       & \multicolumn{1}{c|}{4.5}  & \multicolumn{1}{c|}{10.0} & \multicolumn{1}{c|}{2.9}  & \multicolumn{1}{c|}{0.5}  & \multicolumn{1}{c|}{5.6}  & 4.5  & \multicolumn{1}{c|}{4.7}  & \multicolumn{1}{c|}{10.5} & \multicolumn{1}{c|}{3.8}  & \multicolumn{1}{c|}{0.1}  & \multicolumn{1}{c|}{5.3}  & 5.3  \\ \hline
            \multicolumn{1}{|c|}{\multirow{2}{*}{two-stage}} & Mask-RCNN            & \multicolumn{1}{c|}{22.5} & \multicolumn{1}{c|}{40.3} & \multicolumn{1}{c|}{23.5} & \multicolumn{1}{c|}{\textbf{20.2}} & \multicolumn{1}{c|}{19.4} & 28.9 & \multicolumn{1}{c|}{20.1} & \multicolumn{1}{c|}{41.4} & \multicolumn{1}{c|}{18.0} & \multicolumn{1}{c|}{\textbf{20.2}} & \multicolumn{1}{c|}{17.4} & 25.4 \\ \cline{2-14} 
            \multicolumn{1}{|c|}{}                           & CMask-RCNN    & \multicolumn{1}{c|}{23.0} & \multicolumn{1}{c|}{38.0} & \multicolumn{1}{c|}{\textbf{26.9}} & \multicolumn{1}{c|}{-}    & \multicolumn{1}{c|}{17.8} & \textbf{33.3} & \multicolumn{1}{c|}{20.1} & \multicolumn{1}{c|}{39.4} & \multicolumn{1}{c|}{\textbf{21.9}} & \multicolumn{1}{c|}{-}    & \multicolumn{1}{c|}{16.9} & 27.5 \\ \hline
            \multicolumn{1}{|c|}{our model}                  & DiffRegFormer & \multicolumn{1}{c|}{\textbf{25.7}} & \multicolumn{1}{c|}{\textbf{46.9}} & \multicolumn{1}{c|}{25.8} & \multicolumn{1}{c|}{13.5} & \multicolumn{1}{c|}{\textbf{23.9}} & 31.6 & \multicolumn{1}{c|}{\textbf{23.4}} & \multicolumn{1}{c|}{\textbf{48.8}} & \multicolumn{1}{c|}{20.2} & \multicolumn{1}{c|}{\textbf{20.2}} & \multicolumn{1}{c|}{\textbf{20.4}} & \textbf{30.1} \\ \hline
        \end{tabular}
		\begin{tabular}{|p{3mm}<{\centering}p{22mm}||p{4.5mm}<{\centering}p{4.5mm}<{\centering}p{4.5mm}<{\centering}p{4.5mm}<{\centering}p{4.5mm}<{\centering}p{4.5mm}<{\centering}|p{4.5mm}<{\centering}p{4.5mm}<{\centering}p{4.5mm}<{\centering}p{4.5mm}<{\centering}p{4.5mm}<{\centering}p{4.5mm}<{\centering}|}
            \hline
            \multicolumn{2}{|c||}{\multirow{2}{*}{}}                                 & \multicolumn{6}{c|}{Bounding Boxes - Tubuli}                                                                                                     & \multicolumn{6}{c|}{Instances - Tubuli}                                                                                                          \\ \cline{3-14} 
            \multicolumn{2}{|c||}{}                                                  & \multicolumn{1}{c|}{AP}   & \multicolumn{1}{c|}{AP$_{50}$} & \multicolumn{1}{c|}{AP$_{75}$} & \multicolumn{1}{c|}{AP$_{\mathrm{S}}$}  & \multicolumn{1}{c|}{AP$_{\mathrm{M}}$}  & AP$_{\mathrm{L}}$  & \multicolumn{1}{c|}{AP}   & \multicolumn{1}{c|}{AP$_{50}$} & \multicolumn{1}{c|}{AP$_{75}$} & \multicolumn{1}{c|}{AP$_{\mathrm{S}}$}  & \multicolumn{1}{c|}{AP$_{\mathrm{M}}$}  & AP$_{\mathrm{L}}$  \\ \hline
            \multicolumn{1}{|c|}{\multirow{2}{*}{one-stage}} & QueryInst\_300       & \multicolumn{1}{c|}{45.8} & \multicolumn{1}{c|}{66.4} & \multicolumn{1}{c|}{51.4} & \multicolumn{1}{c|}{\textbf{21.8}} & \multicolumn{1}{c|}{44.3} & 57.9 & \multicolumn{1}{c|}{41.2} & \multicolumn{1}{c|}{67.3} & \multicolumn{1}{c|}{47.6} & \multicolumn{1}{c|}{13.1} & \multicolumn{1}{c|}{39.0} & 55.8 \\ \cline{2-14} 
            \multicolumn{1}{|c|}{}                           & QueryInst\_500       & \multicolumn{1}{c|}{21.2} & \multicolumn{1}{c|}{37.1} & \multicolumn{1}{c|}{21.7} & \multicolumn{1}{c|}{1.3}  & \multicolumn{1}{c|}{20.3} & 30.1 & \multicolumn{1}{c|}{20.6} & \multicolumn{1}{c|}{37.5} & \multicolumn{1}{c|}{20.7} & \multicolumn{1}{c|}{0.7}  & \multicolumn{1}{c|}{18.1} & 40.5 \\ \hline
            \multicolumn{1}{|c|}{\multirow{2}{*}{two-stage}} & Mask-RCNN            & \multicolumn{1}{c|}{57.5} & \multicolumn{1}{c|}{79.6} & \multicolumn{1}{c|}{66.4} & \multicolumn{1}{c|}{15.8} & \multicolumn{1}{c|}{53.8} & 78.2 & \multicolumn{1}{c|}{51.4} & \multicolumn{1}{c|}{79.9} & \multicolumn{1}{c|}{\textbf{62.3}} & \multicolumn{1}{c|}{12.7} & \multicolumn{1}{c|}{47.3} & \textbf{69.8} \\ \cline{2-14} 
            \multicolumn{1}{|c|}{}                           & CMask-RCNN    & \multicolumn{1}{c|}{58.1} & \multicolumn{1}{c|}{79.9} & \multicolumn{1}{c|}{66.5} & \multicolumn{1}{c|}{15.1} & \multicolumn{1}{c|}{\textbf{54.4}} & \textbf{78.8} & \multicolumn{1}{c|}{51.8} & \multicolumn{1}{c|}{80.0} & \multicolumn{1}{c|}{61.7} & \multicolumn{1}{c|}{13.0} & \multicolumn{1}{c|}{\textbf{47.9}} & 69.6 \\ \hline
            \multicolumn{1}{|c|}{our model}                  & DiffRegFormer & \multicolumn{1}{c|}{\textbf{58.5}} & \multicolumn{1}{c|}{\textbf{80.2}} & \multicolumn{1}{c|}{\textbf{67.9}} & \multicolumn{1}{c|}{19.9} & \multicolumn{1}{c|}{51.0} & 75.4 & \multicolumn{1}{c|}{\textbf{55.8}} & \multicolumn{1}{c|}{\textbf{80.6}} & \multicolumn{1}{c|}{\textbf{62.3}} & \multicolumn{1}{c|}{\textbf{16.6}} & \multicolumn{1}{c|}{45.6} & 64.5 \\ \hline
        \end{tabular}
\end{table*}

\section{Results}\label{sec:results}
This section begins with an overview of the kidney biopsy dataset used in this study. We then make a fair comparison between DiffRegFormer and established end-to-end instance segmentation models specifically designed to handle dense, multi-class, and multi-scale objects at the ROI level. {In addition, in order to include retrospective reflection in our analysis, we have compared our model with a semantic segmentation based method \citep{hermsen2019deep}. The results of this comparison are presented in section 4.2.2 and the tables therein.} All benchmark methods are reproduced utilizing the \textbf{mmdetection} package \citep{chen2019mmdetection} to ensure consistency. Further, we evaluate the performance of our lesion classifier on cropped images of anatomical structures derived from the instance masks predicted by DiffRegFormer. Additionally, we have conducted a complexity analysis of DiffRegFormer to identify an optimal balance between its performance and computational burden. Finally, we present extensive ablation studies on DiffRegFormer that evaluate the impact of various training strategies.  
\subsection{Datasets}
\begin{figure}[!htbp]
	\centering
	\includegraphics[width=0.5\textwidth, keepaspectratio=true]{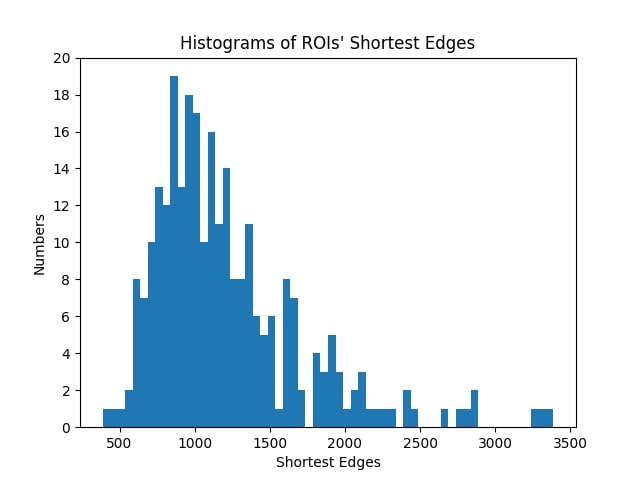}
	\caption{The histogram of the shortest edges of ROIs extracted from WSIs. In our dataset, the shortest ROI length is 387 pixels, and the largest is 4565 pixels.}
	\label{fig:size_distribution}
\end{figure}

The biopsy specimens were prepared according to the Pathology Laboratory Protocol for kidney biopsies. The tissues were collected by a standard procedure of operations: core needle biopsy, formalin fixed, rapidly processed by a routine tissue processor, embedded in paraffin, serially sectioned at $2-4 \mu m$, mounted onto adhesive slides, and stained with Jones silver staining. The dataset comprises $148$ patient biopsy samples (both transplant and native kidneys) collected from the multi-center archive of the Departments of Pathology at LUMC, AUMC, UMCU in the Netherlands or the archive of the Department of Nephrology at Leuven in Belgium. Before analysis, all biopsies were anonymized by a pathology staff member. Annotation was performed using the \textbf{ASAP} version $1.9$ for Windows\footnote{https://computationalpathologygroup.github.io/ASAP/} and our custom software \textbf{Slidescape}\footnote{https://github.com/amspath/slidescape}. The $148$ Jones-stained renal WSIs were acquired from scanners at a resolution of $\mathrm{PPM}=2$ in BIG-TIFF format\footnote{https://www.awaresystems.be/imaging/tiff/bigtiff.html}. Openslide\footnote{https://openslide.org/} was used to extract $303$ ROIs at level $0$ of the BIG-TIFF images. Table \ref{config_dataset} outlines the dataset composition, and Figure \ref{fig:size_distribution} illustrates the distribution of ROI short sizes. Furthermore, LUMC $115$ PAS-stained renal WSIs were used only to test stain transfer. In particular, ROIs from the same WSIs were exclusively allocated to either training or validation. To train the lesion classifier, $21,889$ patches of individual objects were extracted from $303$ ROIs, including $916$ glomeruli patches with binary lesion labels ($382$ with sclerosis and $534$ normal) and $20,973$ tubuli patches with binary labels ($10,647$ with atrophy and $10,326$ normal), as detailed in Figure \ref{config_lesion_dataset}. 
\begin{table}[!ht]
    \centering
    \caption{Origin of datasets digitized with different scanners\label{config_dataset}}
     \begin{minipage}{0.5\textwidth}
        \centering
        \begin{tabular*}{0.9\textwidth}{|p{9mm}<{\centering}|p{23mm}<{\centering}|p{6mm}<{\centering}|p{6mm}<{\centering}|p{6mm}<{\centering}|p{6.5mm}<{\centering}|}
           \hline
           Source & Scanner & WSIs & ROIs & Train & Eval \\
           \hline
           LUMC\footnote{Leiden University Medical Center} & Philips UFS & 65 & 142 & 130 & 12 \\
           \hline
           AUMC & Philips UFS & 42 & 80 & 74 & 6  \\
           \hline
           UMCU & Hamamatsu XR & 25 & 49 & 45 & 4  \\
           \hline
           Leuven & Philips UFS & 16 & 32 & - & 32  \\
           \hline
         \end{tabular*}
    \end{minipage}
 \end{table}

\begin{figure*}[!htbp]
	\centering
	\includegraphics[width=0.95\textwidth, keepaspectratio=true]{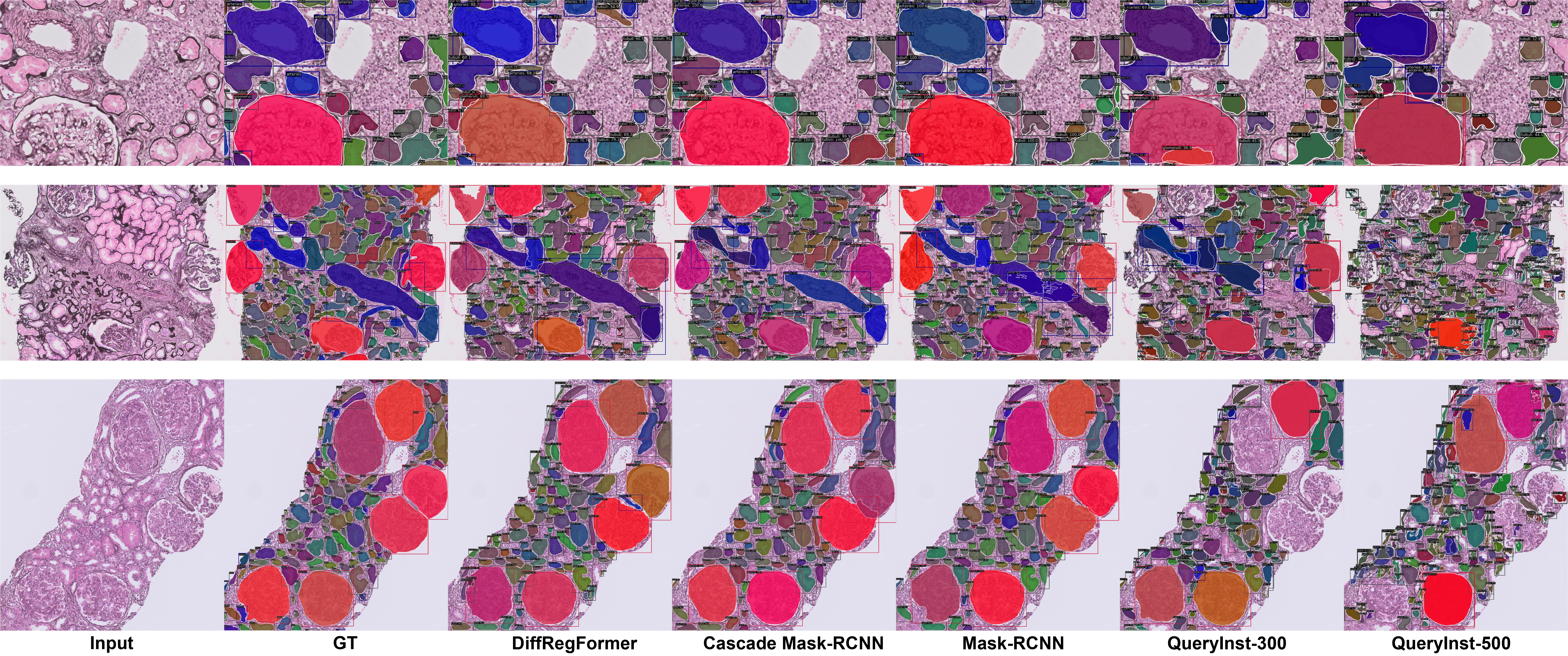}
	\caption{A visual comparison of different instance segment methods on Jones-stained kidney biopsies. It illustrates that our model, i.e., DiffRegFormer, outperforms in detecting objects of diverse size and shape. Despite some instances overlapping, it generates precise instance masks over each region. Mask-RCNN-based models struggle to localize one large object within a single bounding box as they lack attention mechanisms to capture long-range dependencies. QueryInst with {static} queries fails to process dense objects since it fails to detect objects of each class within one ROI.}
	\label{fig:visualization_jones}
    \vspace{0.5em}
	\includegraphics[width=0.95\textwidth, keepaspectratio=true]{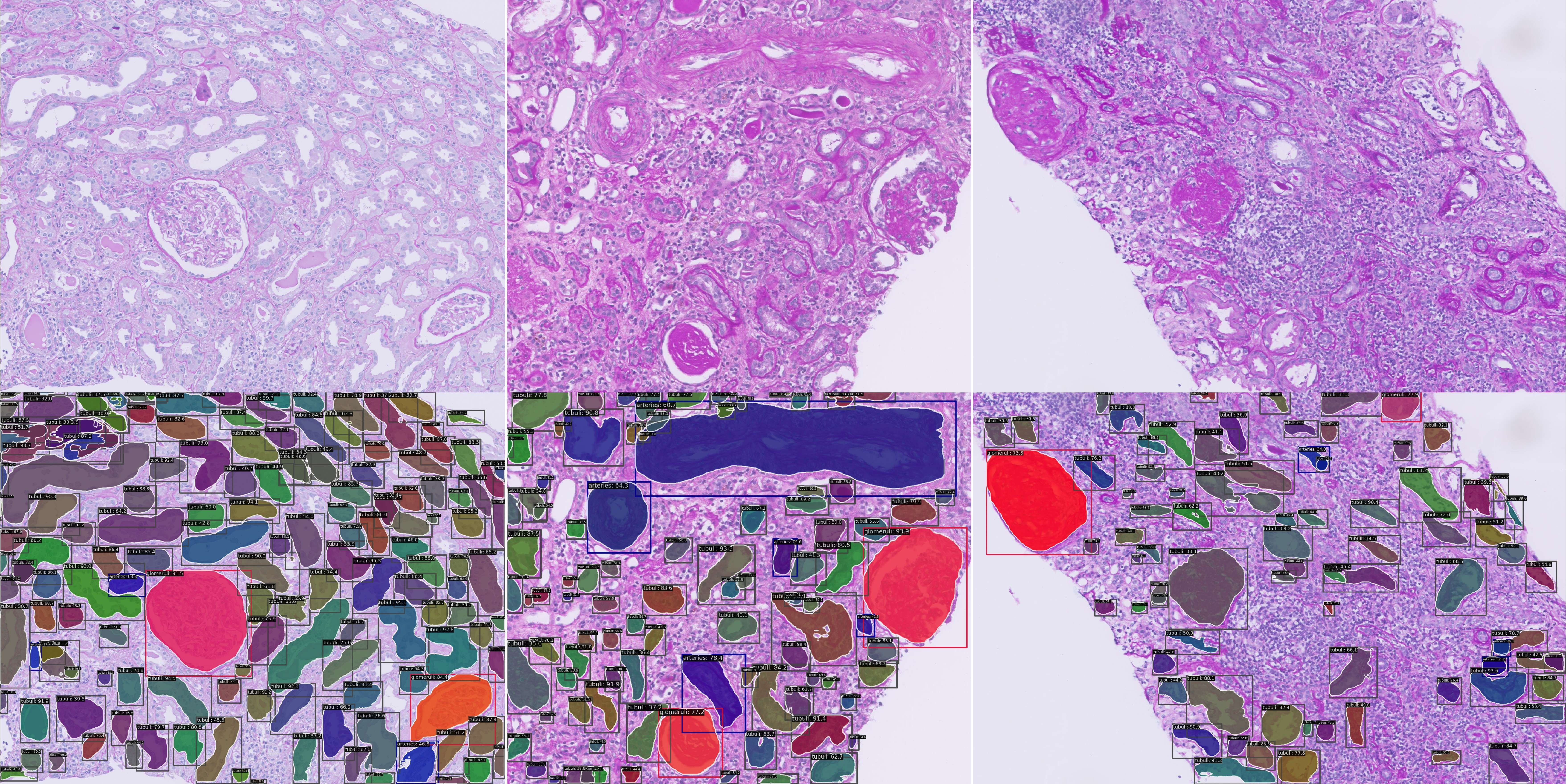}
	\caption{A visualization of results of our model (DiffRegFormer) on the PAS-stained kidney biopsies with increasing degrees of fibrosis and inflammation (from left to right), along with variations in stain intensity.}
	\label{fig:visualization_pas}

    \includegraphics[height=5cm, keepaspectratio=true]{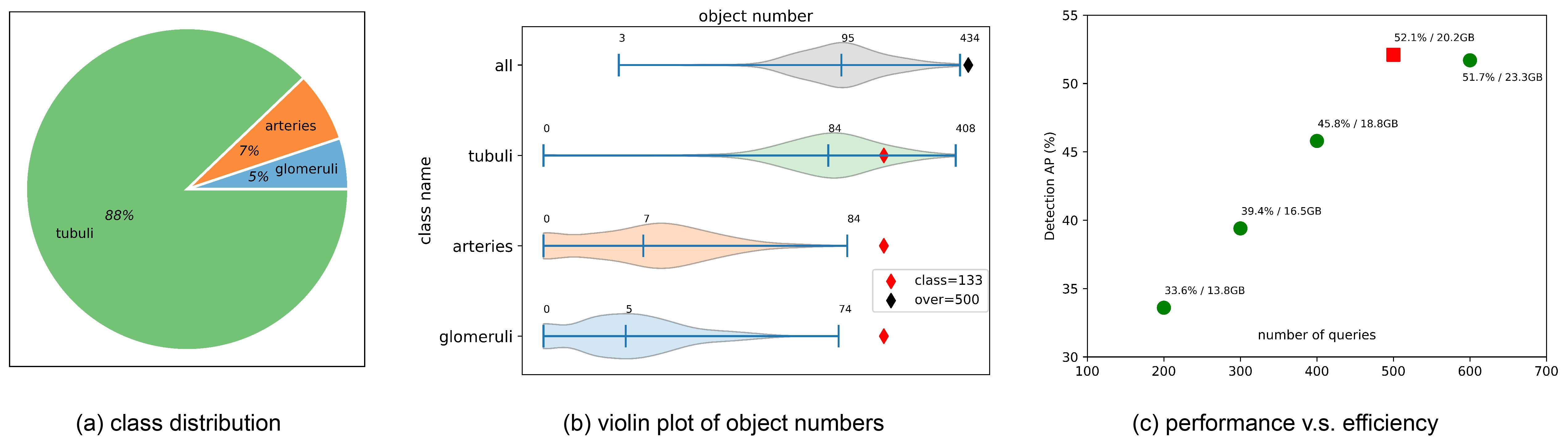}
    \caption{Complexity Analysis on Jones-stained renal biopsy dataset. {Figure (a) is the sample occupation across our dataset. Figure (b) is a violin plot showing the sample number distribution per class. Figure (c) shows trade-offs between performance and efficiency for different experiment settings. The red square is the optimal setting, while the green circles are the sub-optimal settings.}}
    \label{fig:complexity_analysis}
\end{figure*}

\begin{figure}[!htbp]
	\centering
    \captionsetup[subfloat]{labelfont=small,textfont=small}
    \subfloat[shared feature + proposals + biased sampling on MSCOCO's image 
    \label{fig:normal_instance_segmentation}]{%
         \centering
         \includegraphics[width=0.5\textwidth, height=3.5cm]{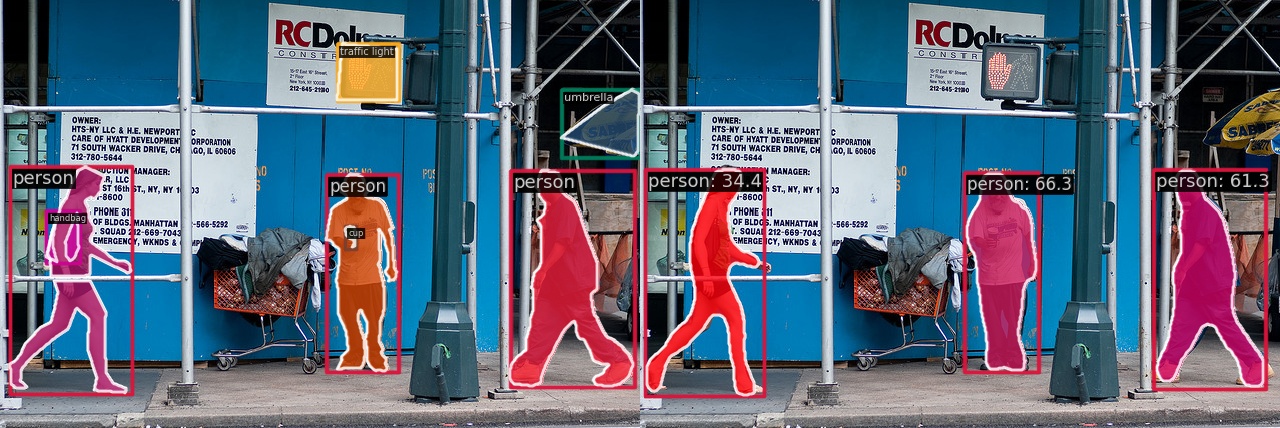}}
     \hfill
     \subfloat[shared feature + ground-truth + biased sampling on a ROI \label{fig:dense_instance_segmentation_shared_feature_plus_gt_plus_biased}]{%
         \centering
         \includegraphics[width=0.5\textwidth, height=3.5cm]{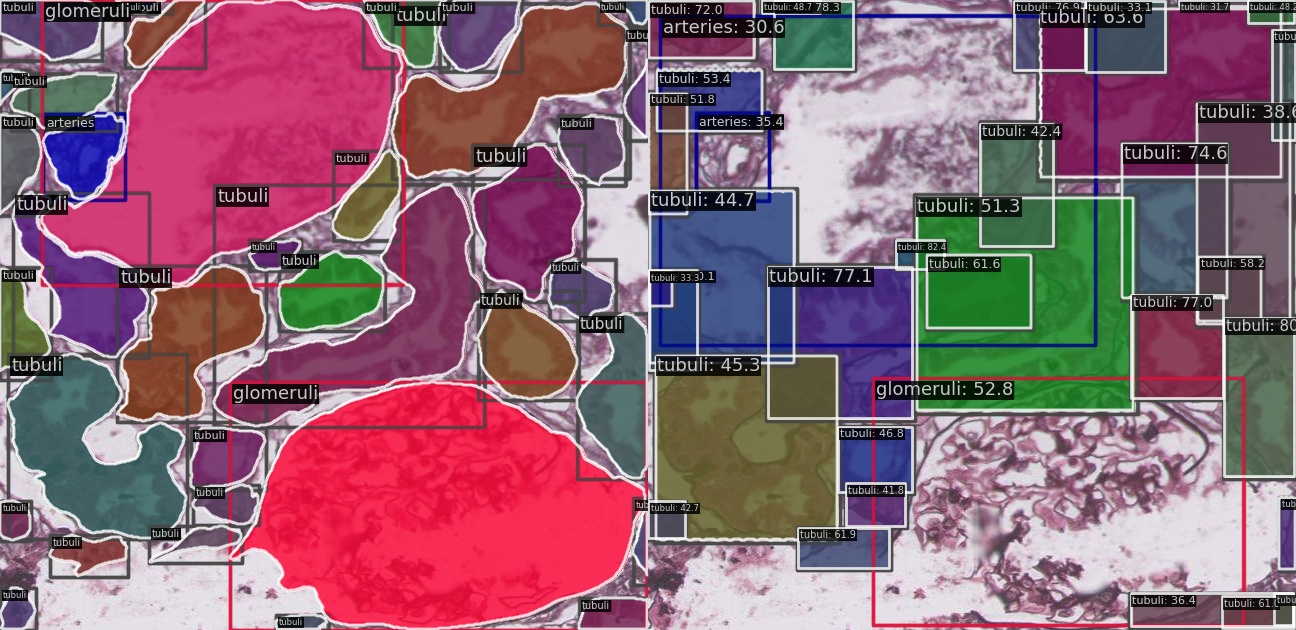}}
     \hfill
     \subfloat[shared feature + ground-truth + unbiased sampling on a ROI\label{fig:dense_instance_segmentation_shared_feature_plus_gt_plus_unbiased}]{%
         \centering
         \includegraphics[width=0.5\textwidth, height=3.5cm]{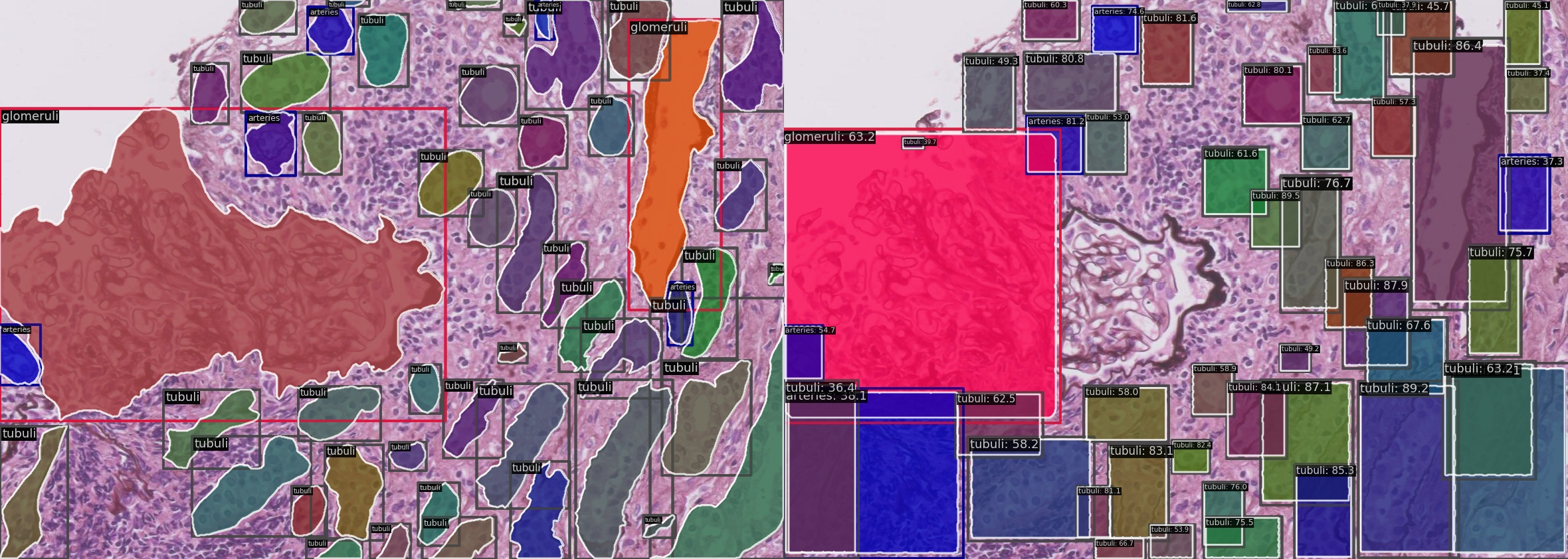}}
     \hfill
     \subfloat[separate feature + proposals + unbiased sampling on a ROI \label{fig:dense_instance_segmentation_separate_feature_plus_proposal_plus_unbiased}]{%
         \centering
         \includegraphics[width=0.5\textwidth, height=3.5cm]{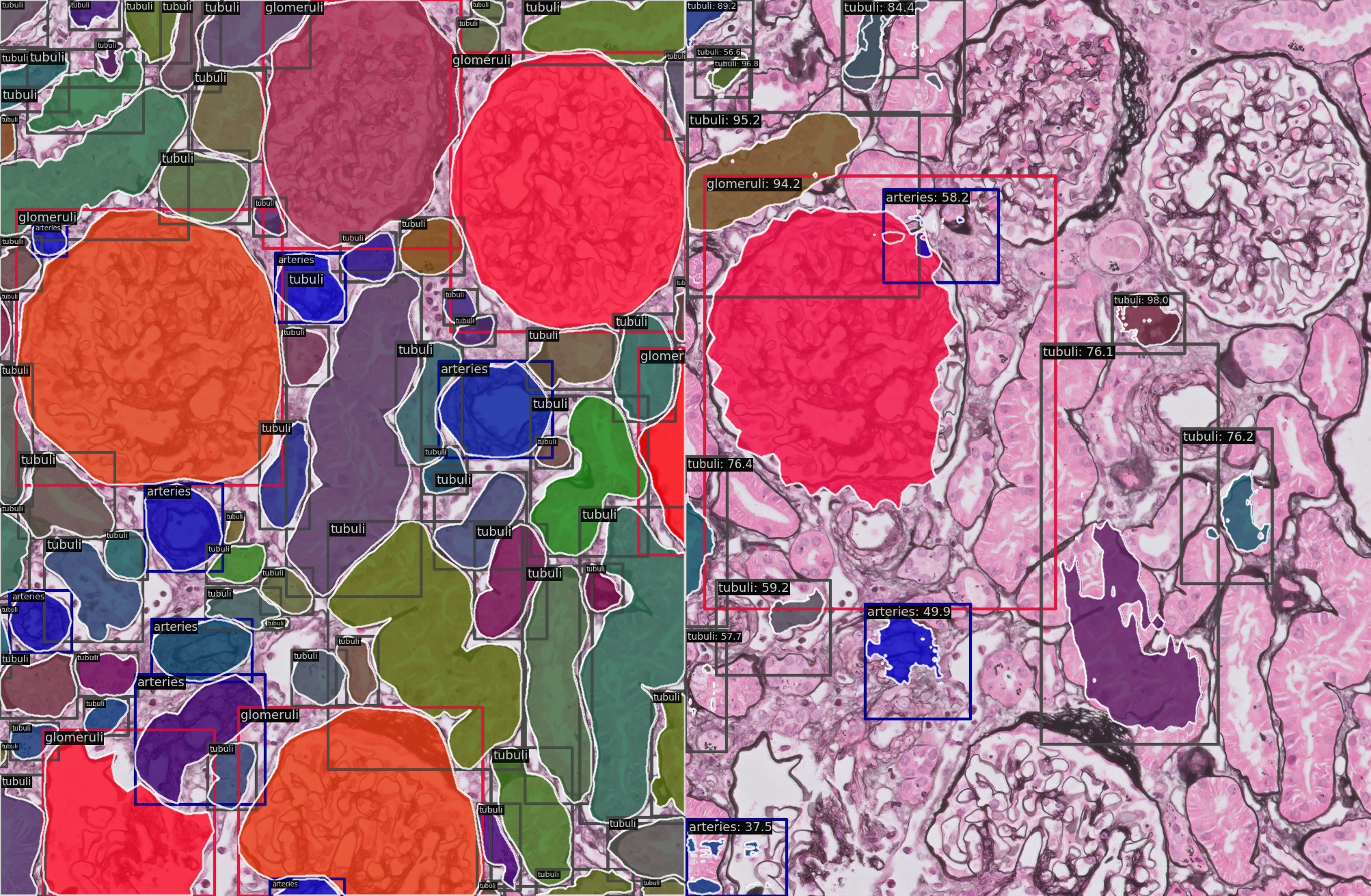}}
     \hfill
     \subfloat[separate feature + ground-truth + biased sampling on a ROI
     \label{fig:dense_instance_segmentation_separate_feature_plus_gt_plus_biased}]{%
         \centering
         \includegraphics[width=0.5\textwidth, height=3.5cm]{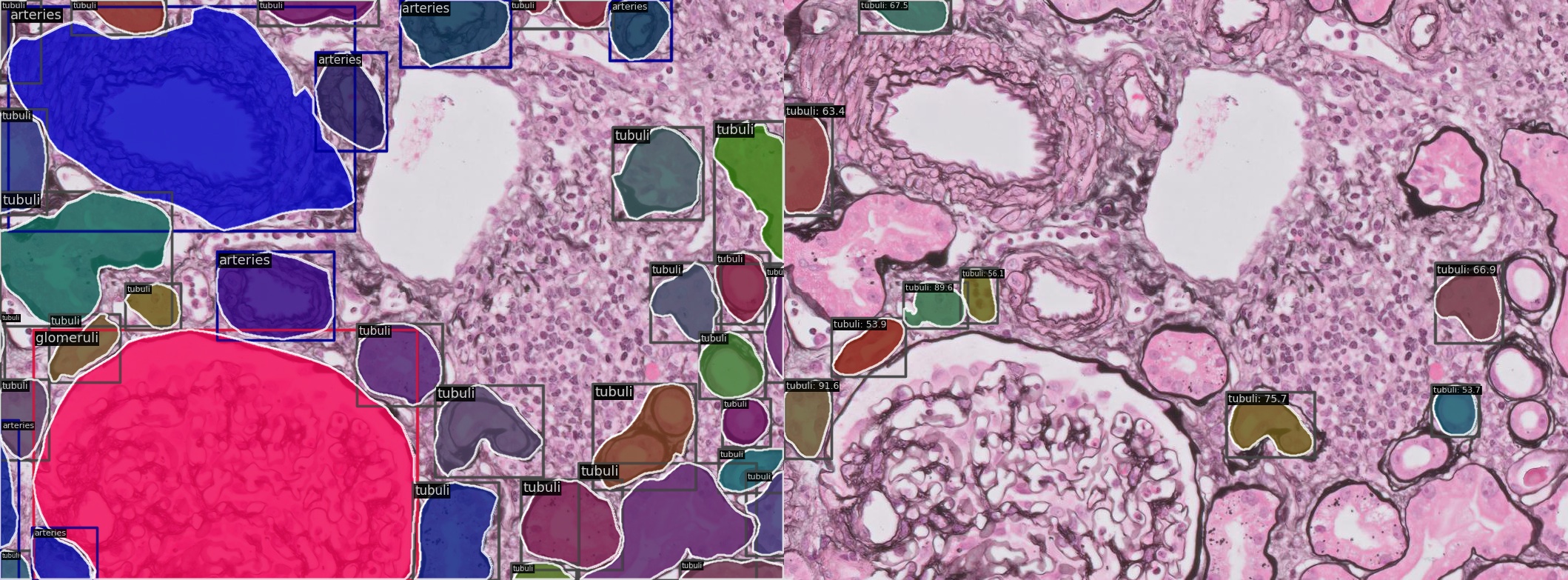}}
     \hfill
	\caption{Ablation study on combinations of different strategies. {The images on the left are ground-truth. The images on the right are our predictions.}}
	\label{fig:distangle_decoders}
\end{figure}

\begin{table*}[!htbp]
    \caption{Evaluation on combinations of different strategies. {The symbol \cmark \ denotes taking one strategy while symbol \xmark \ means taking the converse strategy. The best results are highlighted in bold. \\ \textit{glo}: glomeruli. \textit{art}: arteries. \textit{tub}: tubuli. \textit{all}: overall. \textit{det}: detection. \textit{ins}: instance. "-": not working. \\ \textit{feature}: if use separate feature maps between bbox-decoder and mask-decoder.\\ \textit{boxes}: if use ground-truth boxes in the sampling process.\\ \textit{sample}: if have class balance in the sampling process.
    }}
    \label{ablation_strategies_table}
    \centering
    \begin{tabular*}{\textwidth}{|p{10.5mm}<{\centering}|p{10mm}<{\centering}|p{10.3mm}<{\centering}|p{1.3cm}<{\centering}|p{1.3cm}<{\centering}|p{1.3cm}<{\centering}|p{1.3cm}<{\centering}||p{1.3cm}<{\centering}|p{1.3cm}<{\centering}|p{1.3cm}<{\centering}|p{1.3cm}<{\centering}|}
       \hline
       feature & boxes & sample & glo-det & art-det & tub-det & all-det & glo-ins & art-ins & tub-ins & all-ins \\
       \hline
       \xmark & \xmark & \xmark & - & - & - & - & - & - & - & - \\
       \hline
       \xmark & \xmark & \cmark & - & - & - & - & - & - & - & - \\
       \hline
       \xmark & \cmark & \xmark & 75.8 & 24.9 & 56.4 & 52.4 & - & - & 33.6 & - \\
       \hline
       \xmark & \cmark & \cmark & 76.1 & \textbf{26.2} & 57.6 & 53.3 & 7.3 & 4.8 & 35.2 & 15.8 \\
       \hline
       \cmark & \xmark & \xmark & - & - & - & - & - & - & - & - \\
       \hline
       \cmark & \xmark & \cmark & 8.5 & 4.3 & 13.5 & 8.8 & 6.8 & 4.1 & 12.5 & 7.8 \\
       \hline
       \cmark & \cmark & \xmark & 12.6 & 6.8 & 56.7 & 25.4 & 10.4 & 5.3 & 45.8 & 20.5 \\
       \hline
       \cmark & \cmark & \cmark & \textbf{80.9} & 25.7 & \textbf{58.5} &  \textbf{55.1} & \textbf{74.3} & \textbf{23.4} & \textbf{55.8} & \textbf{51.2} \\
       \hline
     \end{tabular*}
\end{table*}

\begin{table}[!htbp]
    \begin{minipage}{0.48\textwidth}
    \centering
    \caption{Evaluation of detection performances on different types of queries. {$\ast$ best result of each type is highlighted in bold.}}
    \label{ablation_queries_table}
        \begin{tabular*}{\textwidth}{|p{11.8mm}<{\centering}|p{10mm}<{\centering}|p{10mm}<{\centering}|p{10mm}<{\centering}|p{10mm}<{\centering}|p{10mm}<{\centering}|}
           \hline
           \diagbox[innerwidth=11.8mm,font=\footnotesize,innerrightsep=0.5mm,innerleftsep=0.5mm]{mAP}{number} & 200 & 300 & 400 & 500 & 600\\
           \hline
           dynamic & \textbf{33.6} & \textbf{39.4} & \textbf{45.8} & \textbf{52.1}  & \textbf{51.7}  \\
           \hline
           {static} & 23.8 & 38.9 & 37.4  & 34.6 & 31.6  \\
           \hline
        \end{tabular*}
   \end{minipage}
\end{table}
\begin{table}
   \begin{minipage}{0.48\textwidth}
    \centering
    \caption{Lesion dataset configuration and classification Performance}
    \label{config_lesion_dataset}
        \begin{tabular*}{\textwidth}{|p{18.4mm}||p{15.5mm}<{\centering}p{15.5mm}<{\centering}p{15.5mm}<{\centering}p{15.5mm}<{\centering}|}
            \hline
            \multirow{2}{*}{} & \multicolumn{2}{c||}{tubuli}                                & \multicolumn{2}{c|}{glomeruli}         \\ \cline{2-5} 
                              & \multicolumn{1}{c|}{normal} & \multicolumn{1}{c||}{atrophy} & \multicolumn{1}{c|}{normal} & sclerosis \\ \hline
            train             & \multicolumn{1}{c|}{10326}   & \multicolumn{1}{c||}{10647}  & \multicolumn{1}{c|}{534}       & 382    \\ \hline
            eval              & \multicolumn{1}{c|}{906}     & \multicolumn{1}{c||}{568}    & \multicolumn{1}{c|}{51}        & 22     \\ \hline
            per\_precision    & \multicolumn{2}{c||}{85.1}                                  & \multicolumn{2}{c|}{93.2}               \\ \hline
            per\_recall       & \multicolumn{2}{c||}{40.5}                                  & \multicolumn{2}{c|}{88.7}               \\ \hline\hline
            all\_precision    & \multicolumn{4}{c|}{89.2}                                                                           \\ \hline
            all\_recall       & \multicolumn{4}{c|}{64.6}                                                                            \\ \hline
        \end{tabular*}
    \end{minipage}
\end{table}

\begin{figure*}[!htbp]
	\centering
	\includegraphics[height=0.4\textwidth, keepaspectratio=true]{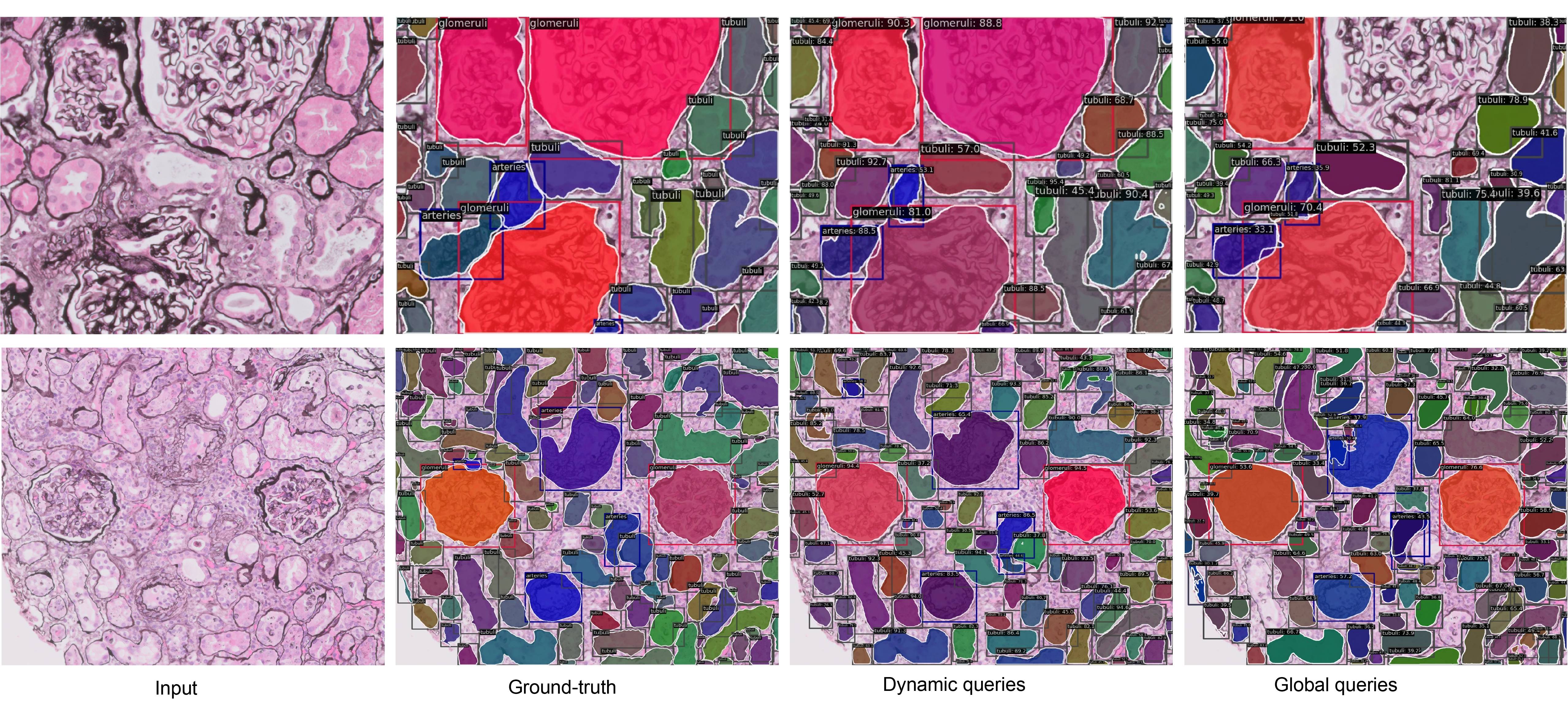}
	\caption{A visual comparison from an ablation study on different query types. Dynamic queries outperform {static} queries on detection. It should be noted that the {static} queries fail to detect  objects of all classes within one ROI.}
	\label{fig:ablation_queries}
\end{figure*}

\begin{figure*}[!htbp]
	\centering
	\includegraphics[height=0.7\textwidth, keepaspectratio=true]{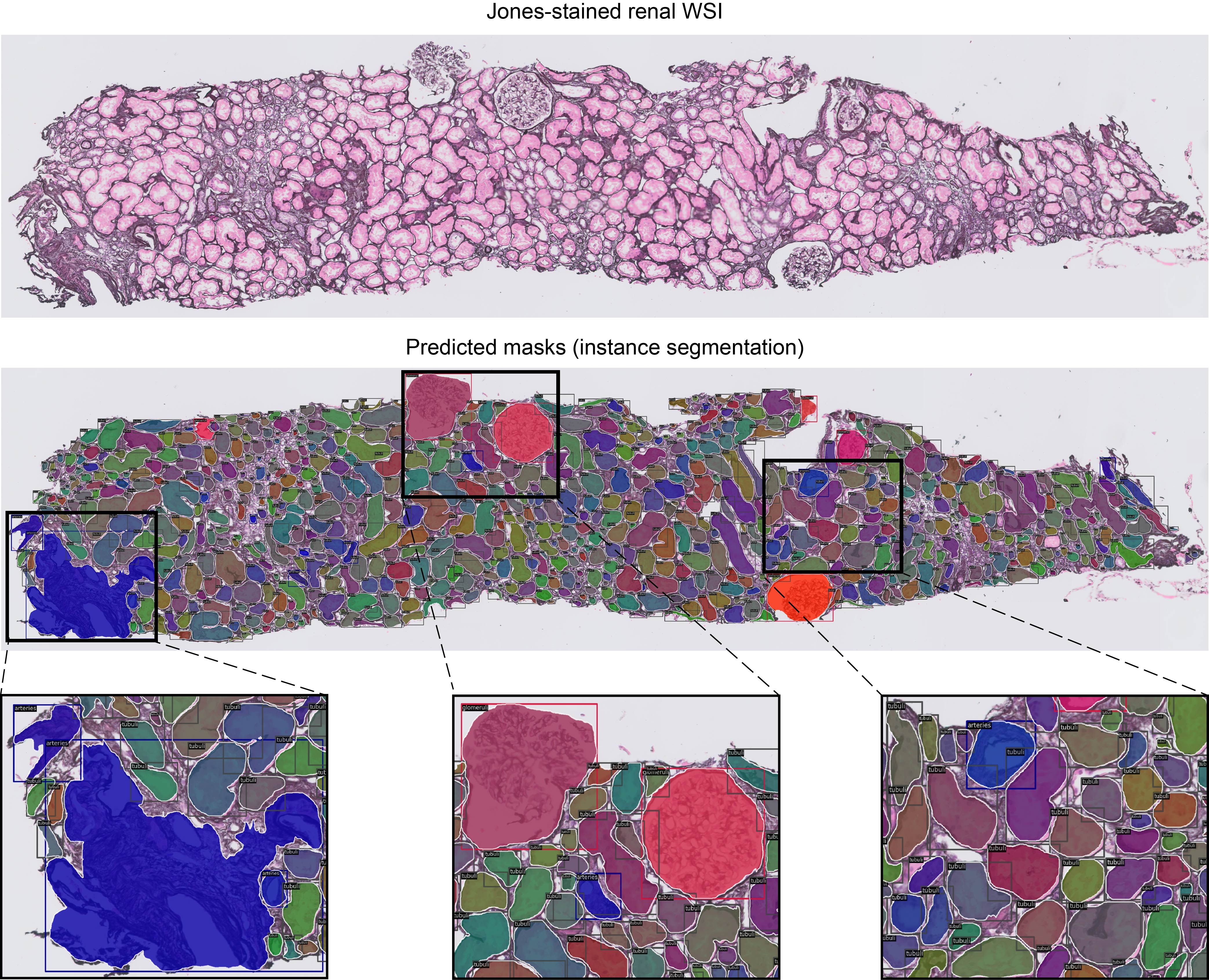}
	\caption{A representative visualization of results from DiffRegFormer stitches showing predicted dense instance objects from a WSI of a Jones-stained renal biopsy.}
	\label{fig:wsi_example}
\end{figure*}

\begin{figure*}[!ht]
	\centering
	\includegraphics[width=0.95\textwidth, keepaspectratio=true]{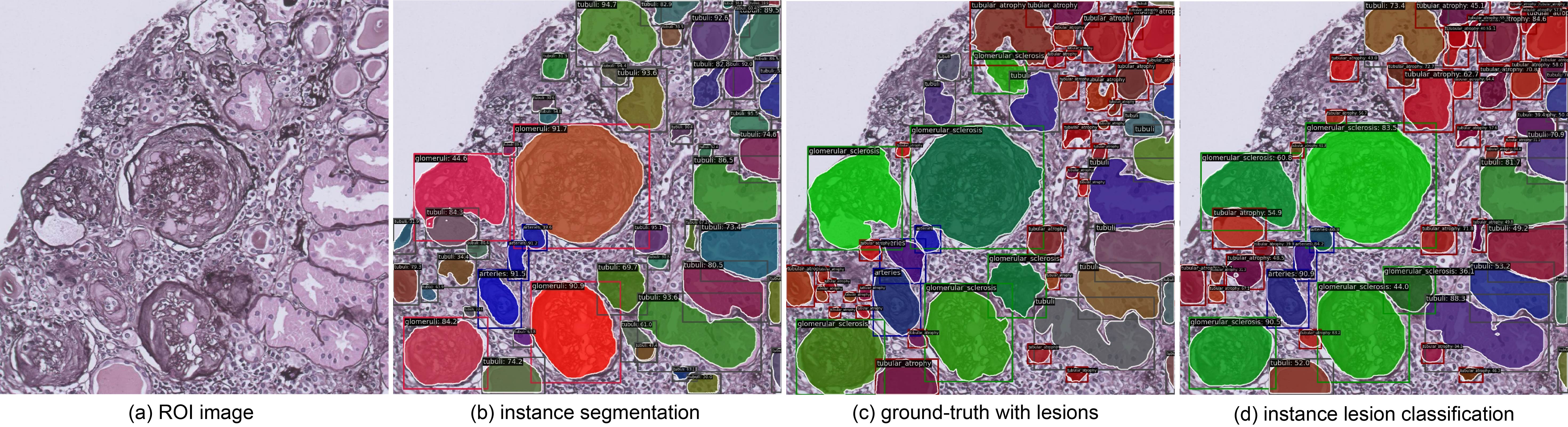}
	\caption{A visualization of results of the lesion classifier predicting the probability of class-wise lesions on an ROI. Combined with dense instance segmentation, our model can generate per-instance lesion identification.}
	\label{fig:quantitive_scoring}
\end{figure*}

\subsection{Experiments}
In our experimental analysis, we conducted a fair comparison of our proposed DiffRegFormer against well-established end-to-end instance segmentation models: Mask R-CNN \citep{he2017mask}, Cascade Mask R-CNN \citep{cai2019cascade}, and QueryInst \citep{fang2021instances}, specifically designed to simultaneously handle dense, multi-class, multi-scale objects at the ROI level. All the models under comparison employed a ResNet-$50$ \citep{he2015deep} backbone pre-trained on the ImageNet \citep{deng2009imagenet} and were trained on $249$ ROIs over $40000$ iterations. We utilized the \textbf{mmdetection} data augmentation pipeline: each ROI was first subjected to \textit{RandomFlip} with a $0.5$ probability, followed by random selection between \textit{RandomResize} or \textit{RandomCrop} also with a probability $0.5$. \textit{RandomResize} adjusted the ROI to a size with the shortest edge between $480$ and $1333$ pixels, while \textit{RandomCrop} extracted patches with the shortest edge between $384$ and $600$ pixels. Finally, pixel values were normalized to $(-1, 1)$. As \textbf{mmdetection} supports training with variable image sizes, all models were allowed to train on objects at multiple scales. \par
For feature extraction, we used $256$ channels for queries. During both training and inference, DiffRegFormer employed $\textbf{500}$ dynamic queries, while QueryInst used $300$ and $500$ {static} queries, respectively. Cascade Mask-RCNN, QueryInst, and our method utilized $6$ cascading stages. Table \ref{eval_tab_overall} presents performance comparisons for object detection w.r.t bounding boxes and instance segmentation w.r.t masks, respectively. For all experiments, DiffRegFormer did not utilize refinement during inference (i.e., iteration set to 1). Notably, Mask-RCNN \citep{he2017mask} and cascade Mask-RCNN \citep{cai2019cascade} are two-stage methods employing RPN networks instead of dynamic queries. QueryInst is a one-stage method with {static} queries. In contrast, DiffRegFormer is categorized as a one-stage method with dynamic queries. \par
\textbf{Detection}. DiffRegFormer achieves an Average Precision (AP) of 52.1\% for detection with a ResNet-50 backbone, thereby surpassing competitors such as QueryInst, Mask R-CNN, and Cascade Mask R-CNN by significant margins. We observe that the iterative refinement hinders detection performance for small objects while benefiting the detection of medium to large objects. It suggests additional operations to mitigate information loss for small objects during the refinement process. Furthermore, {static} queries are less effective for handling dense objects, particularly when the number of queries significantly exceeds the number of feature channels ($300$ or $500$ compared to $256$), leading to a marked decline in detection performance. \par
\textbf{Instance segmentation}. DiffRegFormer achieves a 46.8\% AP for instance segmentation and thus outperforms its counterparts. Moreover, applying iterative refinement to bounding boxes significantly impacts the instance segmentation outcomes. This effect can be attributed to the binary segmentation process within the boxes, where the quality of the instance masks is inherently linked to the recall and precision rates of the bounding boxes. \par
Table \ref{eval_tab_percat} presents the performance per object class. Our model demonstrates an overall superiority in detecting all anatomical structures compared to other methods. Additionally, it performs better in the instance segmentation of glomeruli and arteries while keeping a slight advantage in tubuli segmentation. These findings indicate the ability of our model to process dense, multi-class objects with diverse scales at the ROI level. However, for each model, we observe a performance bottleneck in the processing of arteries, posing a limitation for clinical application due to the significant size variation within the dataset. For instance, the largest arteries are two orders of magnitude larger than the smallest. This fact suggests that more sophisticated attention mechanisms are needed to capture global contextual information for these extreme variations in scale. \par
Figure \ref{fig:visualization_jones} demonstrates DiffRegFormer's capability to detect objects at various scales effectively. It accurately generates instance masks within bounding boxes, demonstrating precise localization of larger arteries ({see} the second row). In contrast, Cascade Mask R-CNN has difficulty locating larger objects within a single bounding box due to the lack of an attention mechanism for capturing long-range dependencies. Furthermore, Mask-RCNN tends to generate multiple detection results for a single object({see} the third row) due to the absence of cascade refinement. These observations suggest that iterative bounding box refinement during training significantly reduces multiple predictions for a single object. At the same time, the attention mechanism with dynamic queries processes large-scale objects. Nevertheless, QueryInst, with its reliance on {static} queries, may struggle to effectively detect all instances in datasets with significant size and shape variations. QueryInst-$300$ tends to miss many structures, while QueryInst-$500$ can even degenerate into generating multiple false predictions for a single object. Furthermore, RCNN-based models share a limitation: a single pixel may be allocated to multiple instance masks due to pixel-wise predictions within each independent ROI. It inherently prevents enforcing the spatial exclusivity of pixels to instances. Additionally, Figure \ref{fig:visualization_pas} depicts DiffRegFormer's potential for domain transfer, as it generates reasonable results on PAS-stained images while being trained exclusively on Jones' silver-stained images. Moreover, Figure \ref{fig:wsi_example} highlights that DiffRegFormer, combined with sliding window and stitching techniques, can be effectively applied to WSIs, further demonstrating its effectiveness. \par
\begin{figure*}[!thbp]
  \centering
  \includegraphics[width=0.95\textwidth, keepaspectratio=true]{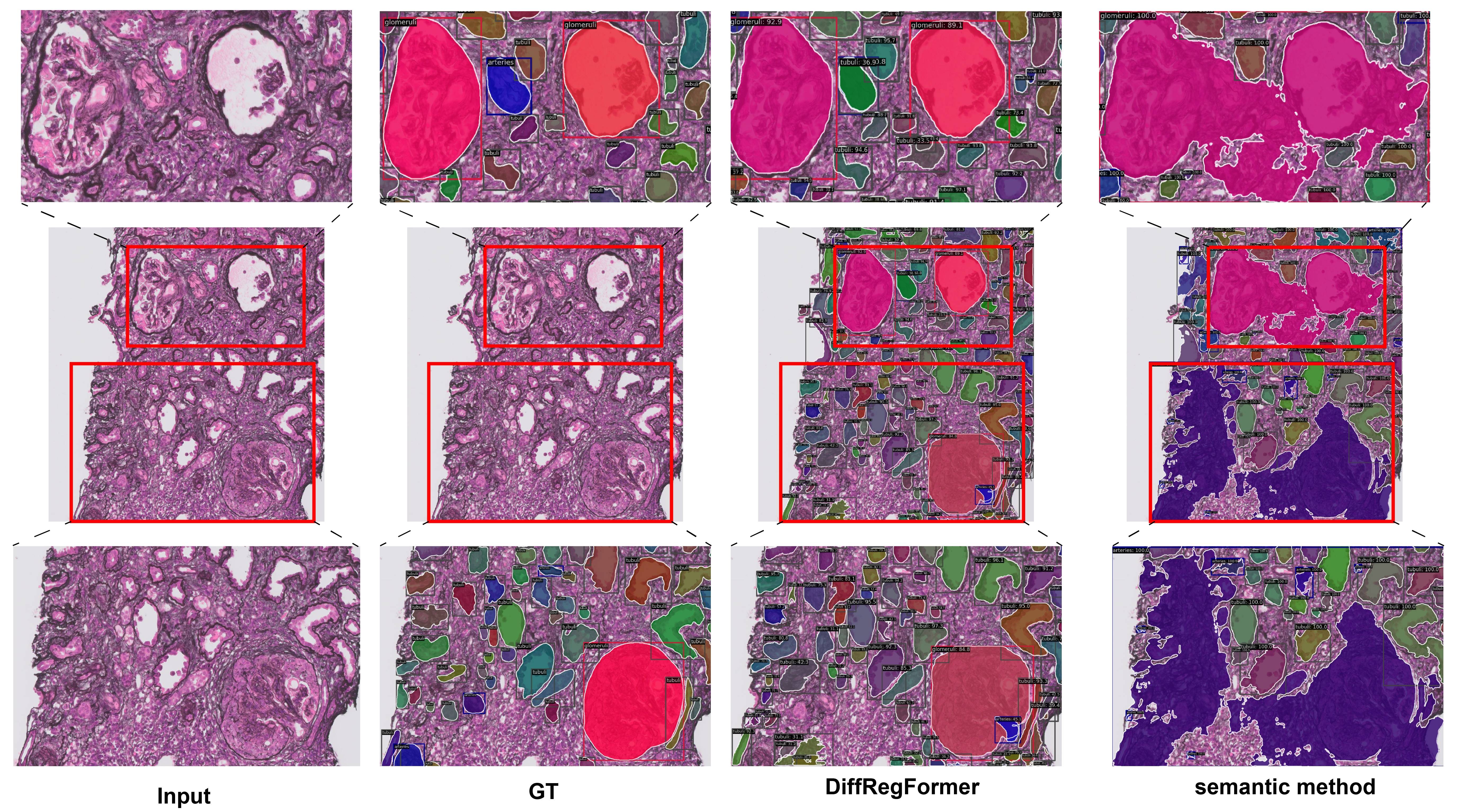}
  \caption{A visual comparison between DiffRegFormer and the semantic method. The upper row represents a zoom of the upper-box in the middle row. The lower row represents a zoom of the lower-box in the middle row. The quality of the result w.r.t. the ground truth labels can be convincingly assessed in favor of DiffRegFormer.\label{fig:vis_compare_semantic}}
\end{figure*}

\textbf{Lesion classification}. The data augmentation pipeline for training the lesion classifier is relatively straightforward. We set the batch size as $96$ for the input croppings, where $32$ patches for each class. Each cropping first undergoes \textit{RandomFlip} with a probability of $0.5$, followed by resizing to $256\times 256$ pixels. Finally, pixel values are normalized to the $(-1, 1)$ range. As shown in Table \ref{config_lesion_dataset}, the classifier achieves an overall precision of $89.2\%$ and a recall of $64.6\%$. More specifically, per-class precision for atrophic tubuli is $85.1\%$, and $93.2\%$ for sclerotic glomeruli. In addition, per-class recall is $40.5\%$ for atrophic tubuli and $88.7\%$ for sclerotic glomeruli. Since the lesion classifier operates on croppings of the anatomical structure generated by DiffRegFormer during inference, its recall rate heavily depends on the performance of our dense instance segmentation model. However, the high precision indicates that lesion classification on correctly cropped structures is very accurate. In summary, the enhancement of the performance of DiffRegFormer is expected to improve the recall of the lesion classifier further. Figure \ref{fig:quantitive_scoring} visualizes the mapping of lesion predictions onto the instance segmentation results. It reveals that our model can be a solid foundation for further clinical diagnosis. 
{\subsubsection{ Metrics for Instance Segmentation}}
{We have computed a set of commonly used metrics to assess the performance of our method in a broader scope. These are DICE, MASD (mean average surface distance), FLOPS, inference speed, and amount of parameters. In Table \ref{tab_extra_metric}, we can observe from the values of DICE and MASD that our method outperforms other approaches on segmentation performance. The resource efficiency is, however, moderate. The main cause of resource cost is the attention mechanism, as seen in QueryInst and DiffRegFormer. To that end, it is beneficial to adopt more efficient attention algorithms to accelerate the inference speed and lessen resource burden, tailoring for the practical clinical agenda. In addition, the arteries that vary across multiple scales make it difficult to obtain high-quality instance masks. All methods have illustrated a large average surface distance between the predicted and ground-truth boundaries.}   
\begin{table}[!htbp]
\caption{Overall Metrics for method assessment. \textit{T} stands for trillion floating-point operations per second. \textit{sec} means second in time. \textit{Para} refers to the number of parameters where \textit{M} denotes million.}\label{tab_extra_metric}
\small
\begin{tabular}{
|p{0.5mm}<{\centering}p{19mm}<{\centering}|p{5.5mm}<{\centering}|p{5.5mm}<{\centering}|p{5.5mm}<{\centering}|p{5.5mm}<{\centering}|p{5.5mm}<{\centering}|
}
\hline
\multicolumn{2}{|c|}{}                                            & {\scriptsize MASD}  & {\scriptsize DICE} & \begin{tabular}[c]{@{}c@{}}{\scriptsize FLOPS}\\ {\scriptsize (T)}\end{tabular} & \begin{tabular}[c]{@{}c@{}}{\scriptsize Speed}\\ {\scriptsize (sec)}\end{tabular} & \begin{tabular}[c]{@{}c@{}}{\scriptsize Para}\\ {\scriptsize (M)}\end{tabular} \\ \hline
\multicolumn{1}{|c|}{\multirow{2}{*}{one-stage}} & QueryInst\_300 & 7.49  & 42.6 & 0.41  & 1.425                                                 & 173       \\ \cline{2-7} 
\multicolumn{1}{|c|}{}                           & QueryInst\_500 & 10.95 & 35.9 & 0.55  & 1.647                                                 & 246       \\ \hline
\multicolumn{1}{|c|}{\multirow{2}{*}{two-stage}} & Mask-RCNN      & 4.78  & 50.7 & 0.20 & \textbf{0.864}                                                 & \textbf{43.98}    \\ \cline{2-7} 
\multicolumn{1}{|c|}{}                           & CMask-RCNN     & 3.87  & 52.1 & \textbf{1.72} & 1.156                                                 & 77.03    \\ \hline
\multicolumn{1}{|c|}{our model}                  & DiffRegFormer  & \textbf{3.64}  & \textbf{56.6} & 0.55 & 1.324                                                 & 140       \\ \hline
\end{tabular}
\end{table}
{\subsubsection{Comparison with Segmentation Based Method}}
{In order to supplement our analysis in terms of completeness, we have done a retrospective analysis reproducing a semantic segmentation based instance segmentation framework \citep{hermsen2019deep}. We denote this method as a semantic method and compare its results with ours. In Table \ref{eval_tab_overall_extra}, an overall comparison is depicted. Our method outperforms the semantic method in prediction quality at the cost of more inference time and computational resources. In Table \ref{eval_tab_percat_extra}, the superiority of our method per class is depicted. Furthermore, Figure \ref{fig:vis_compare_semantic} illustrates that the semantic method is vulnerable to the segmentation quality of the auxiliary boundary and interstitium class, which is insufficient in separating objects tightly touching. Such touching objects are typical in kidney tissue. Our method, on the contrary, can detect and segment each object.}
\begin{table*}[!htpb]
\caption{Overall evaluation result. The semantic method is an instance segmentation framework that combines semantic segmentation with a post-processing operation. We have reproduced it following the literature \citep{hermsen2019deep}. \textit{AP} denotes average precision.\label{eval_tab_overall_extra}}
\resizebox{1.0\textwidth}{!}{
\begin{tabular}{|l|ccc|ccc|ccccc|}
\hline
\multirow{2}{*}{}                     & \multicolumn{3}{c|}{Bounding Boxes}                          & \multicolumn{3}{c|}{Instance}                                & \multicolumn{5}{c|}{Extra Metrics}                                                                                                                \\ \cline{2-12} 
                                      & \multicolumn{1}{c|}{AP}   & \multicolumn{1}{c|}{AP$_{50}$} & AP$_{75}$ & \multicolumn{1}{c|}{AP}   & \multicolumn{1}{c|}{AP$_{50}$} & AP$_{75}$ & \multicolumn{1}{c|}{DICE} & \multicolumn{1}{c|}{MASD} & \multicolumn{1}{c|}{\begin{tabular}[c]{@{}c@{}}{FLOPS}\\ {(T)}\end{tabular}}  & \multicolumn{1}{c|}{\begin{tabular}[c]{@{}c@{}}Speed\\ (sec)\end{tabular}} & \begin{tabular}[c]{@{}c@{}}Para\\ (M)\end{tabular} \\ \hline
\multicolumn{1}{|c|}{DiffRegFormer}   & \multicolumn{1}{c|}{\textbf{52.1}} & \multicolumn{1}{c|}{\textbf{71.1}} & \textbf{57.7} & \multicolumn{1}{c|}{\textbf{46.8}} & \multicolumn{1}{c|}{\textbf{71.6}} & \textbf{52.8} & \multicolumn{1}{c|}{\textbf{56.6}} & \multicolumn{1}{c|}{\textbf{3.87}} & \multicolumn{1}{c|}{0.55} & \multicolumn{1}{c|}{1.324}                                                 & 140       \\ \hline
\multicolumn{1}{|c|}{semantic method} & \multicolumn{1}{c|}{44.6} & \multicolumn{1}{c|}{64.8} & 46.7 & \multicolumn{1}{c|}{40.3} & \multicolumn{1}{c|}{65.3} & 44.6 & \multicolumn{1}{c|}{45.4} & \multicolumn{1}{c|}{7.86} & \multicolumn{1}{c|}{\textbf{1.24}} & \multicolumn{1}{c|}{\textbf{0.760}}                                                 & \textbf{32.8}      \\ \hline
\end{tabular}
}
\end{table*}
\begin{table*}[!htbp]
    \centering
	\caption{Evaluation results per object class. \textit{AP} denotes average precision.\label{eval_tab_percat_extra}}
		\begin{tabular}{|l|ccc|ccc|cc|}
        \hline
        \multirow{2}{*}{}                     & \multicolumn{3}{c|}{Bounding Boxes - glo}                          & \multicolumn{3}{c|}{Instance - glo}                                & \multicolumn{2}{c|}{Extra Metrics} \\ \cline{2-9} 
                                              & \multicolumn{1}{c|}{AP}   & \multicolumn{1}{c|}{AP$_{50}$} & AP$_{70}$ & \multicolumn{1}{c|}{AP}   & \multicolumn{1}{c|}{AP$_{50}$} & AP$_{70}$ & \multicolumn{1}{c|}{DICE} & MASD        \\ \hline
        \multicolumn{1}{|c|}{DiffRegFormer}   & \multicolumn{1}{c|}{80.9} & \multicolumn{1}{c|}{94.2} & 89.8 & \multicolumn{1}{c|}{74.3} & \multicolumn{1}{c|}{94.2} & 80.9 & \multicolumn{1}{c|}{78.8}  & 1.21      \\ \hline
        \multicolumn{1}{|c|}{semantic method} & \multicolumn{1}{c|}{75.6} & \multicolumn{1}{c|}{82.4} & 81.2 & \multicolumn{1}{c|}{68.3} & \multicolumn{1}{c|}{84.3} & 70.1 & \multicolumn{1}{c|}{70.6} &  8.68      \\ \hline
        \end{tabular}
		\begin{tabular}{|l|ccc|ccc|cc|}
        \hline
        \multirow{2}{*}{}                     & \multicolumn{3}{c|}{Bounding Boxes - art}                          & \multicolumn{3}{c|}{Instance - art}                                & \multicolumn{2}{c|}{Extra Metrics} \\ \cline{2-9} 
                                              & \multicolumn{1}{c|}{AP}   & \multicolumn{1}{c|}{AP$_{50}$} & AP$_{75}$ & \multicolumn{1}{c|}{AP}   & \multicolumn{1}{c|}{AP$_{50}$} & AP$_{75}$ & \multicolumn{1}{c|}{DICE} & MASD         \\ \hline
        \multicolumn{1}{|c|}{DiffRegFormer}   & \multicolumn{1}{c|}{25.7} & \multicolumn{1}{c|}{46.9} & 25.8 & \multicolumn{1}{c|}{23.4} & \multicolumn{1}{c|}{48.8} & 20.2 & \multicolumn{1}{c|}{30.2} & 25.94       \\ \hline
        \multicolumn{1}{|c|}{semantic method} & \multicolumn{1}{c|}{10.3} & \multicolumn{1}{c|}{25.8} & 13.6 & \multicolumn{1}{c|}{11.5} & \multicolumn{1}{c|}{24.9} & 10.3 & \multicolumn{1}{c|}{16.3}  & 49.14      \\ \hline
        \end{tabular}
		\begin{tabular}{|l|ccc|ccc|cc|}
        \hline
        \multirow{2}{*}{}                     & \multicolumn{3}{c|}{Bounding Boxes - tub}                          & \multicolumn{3}{c|}{Instance - tub}                                & \multicolumn{2}{c|}{Extra Metrics} \\ \cline{2-9} 
                                              & \multicolumn{1}{c|}{AP}   & \multicolumn{1}{c|}{AP$_{50}$} & AP$_{75}$ & \multicolumn{1}{c|}{AP}   & \multicolumn{1}{c|}{AP$_{50}$} & AP$_{75}$ & \multicolumn{1}{c|}{DICE} & MASD         \\ \hline
        \multicolumn{1}{|c|}{DiffRegFormer}   & \multicolumn{1}{c|}{58.5} & \multicolumn{1}{c|}{80.2} & 67.9 & \multicolumn{1}{c|}{55.8} & \multicolumn{1}{c|}{80.6} & 62.3 & \multicolumn{1}{c|}{64.8} & 0.89       \\ \hline
        \multicolumn{1}{|c|}{semantic method} & \multicolumn{1}{c|}{43.6} & \multicolumn{1}{c|}{73.5} & 54.6 & \multicolumn{1}{c|}{46.9} & \multicolumn{1}{c|}{75.2} & 52.4 & \multicolumn{1}{c|}{50.5} &  2.16      \\ \hline
        \end{tabular}
\end{table*}
\subsection{Complexity Assessment of Model Training}\label{subsec:complexity_analysis}
We start with analyzing the intrinsic class imbalance within our dataset. Of the total structures ($21,889$ objects), 88\% are tubuli, followed by 7\% arteries (1679 objects), and 5\% glomeruli (1226 objects). This distribution is depicted in Figure \ref{fig:complexity_analysis}a, highlighting the necessity of sampling to include more rare instances, such as arteries and glomeruli, among the dominant tubuli. A violin plot, shown in Figure \ref{fig:complexity_analysis}b, details the distribution of structure counts at the ROI level. On average, ROIs contain $95$ objects, with a maximum of $434$. To accommodate all objects, DiffRegFormer is configured to handle a maximum number exceeding $434$ per image. Figure \ref{fig:complexity_analysis}c depicts the correlation between the increase in computational cost and performance improvement. As expected,  we observe a turning point at $500$ dynamic queries. This configuration yields a favorable trade-off between complexity and efficiency, achieving 52.1\% AP with $20.2$GB GPU memory usage. Specifically, ROIs contain an average of $5$ glomeruli, $7$ arteries, and $84$ tubuli with corresponding maximums of $74$, $84$, and $408$. Setting the maximum number per class to $166$ is another feasible option for positive sample generation, as our sampling approach ensures comprehensive inclusion of rare instances while maintaining the rational proportion of the majority class. In conclusion, selecting a maximum of $500$ objects per image and $166$ per class can effectively balance performance and computational efficiency.
\subsection{Ablation Study}\label{subsec:ablation}
The ablation study focuses on the effects of various training strategies and query types on model performance. By analyzing our results, we can gain insights into the specific characteristics of each strategy. This analysis highlights the significance of our improvements as key contributions to adapting an end-to-end instance segmentation model to datasets with dense objects.
\subsubsection{Training strategies}\label{subsubsec:ablation_train_strategies}
Training an instance segmentation model for datasets with dense objects, such as, in our case, kidney biopsies, presents distinct challenges compared to the common datasets with sparser object distributions like MSCOCO. Instance segmentation models typically employ three strategies: (1) shared features between the bbox-decoder and mask-decoder; (2) selection of positive samples from proposal bounding boxes; and (3) class-imbalanced sampling. However, for dense object segmentation, we propose three alternative strategies: (a) separate features for the bbox and mask decoders; (b) positive sample selection directly from ground-truth boxes; and (c) class-balanced sampling. While the conventional combination of shared features, proposal boxes, and biased sampling is adequate for standard instance segmentation ({see} Figure \ref{fig:normal_instance_segmentation}), only the combination of separate features, ground-truth boxes, and unbiased sampling proves to be successful for dense instance segmentation. Table \ref{ablation_strategies_table} evaluates the combinations above, with conventional strategies marked with \xmark \ and our proposed strategy with \cmark. Notably, only five out of eight possible combinations can train DiffRegFormer. \par
Figure \ref{fig:dense_instance_segmentation_shared_feature_plus_gt_plus_biased} demonstrates that combining shared features, ground-truth boxes, and biased sampling enables object detection across all classes. However, it fails to produce accurate instance masks within these bounding boxes. The observation suggests that the bbox-decoder dominates the shared feature representation, thus hindering the mask decoder's ability to extract pixel-wise information. Consequently, all pixels within detected boxes are classified as tubuli. Moreover, biased sampling impedes the mask decoder's capacity to learn features for rare instances (glomeruli and arteries), resulting in no pixel classification within those boxes. \par
Additionally, Figure \ref{fig:dense_instance_segmentation_shared_feature_plus_gt_plus_unbiased}, shared features combined with ground-truth boxes and unbiased sampling can detect and segment objects across all classes, albeit the degraded mask quality. This degradation is likely due to the bbox-decoder's dominance within the shared feature representation. \par
In contrast,  Figure \ref{fig:dense_instance_segmentation_separate_feature_plus_proposal_plus_unbiased} reveals that using separate features, proposal boxes, and unbiased sampling leads to inaccurate object detection. It is probably due to the accumulation of errors introduced by proposals during the early stages of diffusion model training. \par
Lastly, Figure \ref{fig:dense_instance_segmentation_separate_feature_plus_gt_plus_biased} demonstrates that separate features, ground-truth boxes, and biased sampling result in correct detection and segmentation for only a small proportion of tubuli. This finding further highlights the challenges caused by biased sampling in learning features for rare instances. \par 
In summary, shared feature maps lead to the bbox-decoder dominating feature representation learning, impairing the mask-decoder's ability to extract pixel-wise information necessary for accurate binary instance masks. Biased sampling further hinders the mask-decoder from learning information to segment rare instances (glomeruli and arteries), causing it to be overwhelmed by the majority class (tubuli). Finally, utilizing proposals for sampling instead of ground-truth boxes introduces cumulative errors that disrupt the training for both decoders, preventing the learning of contextual information necessary for long-range correlations via attention mechanisms.
\subsubsection{Queries}
{Static} queries in transformer-based models face challenges in representing dense objects within large-scale datasets due to their static nature. Adequate accommodation to extensive variations in object size and shape necessitates a proportional increase in query count. Alternatively, dynamic queries, generated on-the-fly for each object in a mini-batch, demonstrate superior adaptability to dataset complexity. Theoretically, assigning more {static} queries could achieve comparable performance to dynamic queries. However, our experiments indicate that increasing the number of {static} queries without a corresponding increase in feature channels does not yield improvements. Consequently, employing {static} queries for dense object processing in transformer-based models is significantly less cost-effective than dynamic queries. \par 
Table \ref{ablation_queries_table} delineates our comparative analysis of varying query counts for both dynamic and {static} queries, with the number of feature channels fixed at $256$. We observe that increasing the number of dynamic queries leads to a gain in performance while increasing the number of {static} queries results in performance degradation. Figure \ref{fig:ablation_queries} further illustrates the inherent challenge of achieving a favorable trade-off between complexity and performance with {static} queries. Even with $500$ queries, the model with {static} queries fails to detect structures across all classes consistently. This limitation is not attributable to class imbalance compared to the dynamic query strategy. In conclusion, the dynamic query is a more efficient approach for datasets with dense objects, highlighting their superiority over {static} queries.
\section{Conclusion and Future Work}\label{sec:conclusion}
Our research introduces a novel framework to process large-scale renal WSI datasets with potential changes in lesion combinations. Our model performs better in predicting lesions within dense objects, effectively handling multi-class and multi-scale challenges at the ROI level. We present the first dense instance segmentation module, DiffRegFormer, that seamlessly integrates a diffusion model with a transformer in RCNN-style. This approach leverages the diffusion model to generate refined object boxes, upon which a regional transformer creates dynamic queries that are ultimately transformed into accurate instance masks within the boxes. Furthermore, our lesion classifier accurately predicts class-specific lesions for each cropped structure. The model demonstrates several advantages in handling dense object datasets: (1) its iterative noise-to-box approach eliminates the dependence on prior knowledge about object size and shape, facilitating rapid object localization; (2) dynamic queries, generated on-the-fly for each object in a mini-batch style, ensuring scalability to large datasets; (3) the modular design of the model, with independent plug-in prediction heads, allows for replacement and promotes adaptability. This flexibility enables the efficient reuse of trained components when up-scaling to large-scale datasets. Experimental results on Jones' stained renal biopsies demonstrate that our model surpasses existing benchmarks. \par
The current model needs further exploration to meet the rigorous requirements for clinical applications. Still, it establishes a robust framework for dense, multi-class, multi-scale object recognition at the ROI level, setting a solid foundation for future computational improvements. Potential areas for future research include: {(1) further enhancing the compatibility of diffusion models with transformers like DETR to avoid NMS post-processing.} (2) developing innovative regional feature extractors to generate more informative dynamic queries and mitigate the loss of information for small objects during iterative refinements. (3) integrating spatial constraints into regional mask inference to prevent erroneous assignment of single pixels to multiple instances. {(4) designing lightweight attention mechanisms to lessen the computational burden.}
In conclusion, the approach presented paves the way for efficient and reliable automated assessment as a tool for the workbench of renal pathologists. 
\section{Acknowledgements }\label{sec:acknowledgements }
This work is supported in part by funds from the Dutch Kidney Foundation (17OKG23), the Research Priority Area Human(e) AI at the University of Amsterdam, and the Chinese Scholarship Council (CSC).
\bibliographystyle{model2-names.bst}\biboptions{authoryear}
\bibliography{refs}

\begin{thebibliography}{62}
\expandafter\ifx\csname natexlab\endcsname\relax\def\natexlab#1{#1}\fi
\providecommand{\url}[1]{\texttt{#1}}
\providecommand{\href}[2]{#2}
\providecommand{\path}[1]{#1}
\providecommand{\DOIprefix}{doi:}
\providecommand{\ArXivprefix}{arXiv:}
\providecommand{\URLprefix}{URL: }
\providecommand{\Pubmedprefix}{pmid:}
\providecommand{\doi}[1]{\href{http://dx.doi.org/#1}{\path{#1}}}
\providecommand{\Pubmed}[1]{\href{pmid:#1}{\path{#1}}}
\providecommand{\bibinfo}[2]{#2}
\ifx\xfnm\relax \def\xfnm[#1]{\unskip,\space#1}\fi
\bibitem[{Alnazer et~al.(2021)Alnazer, Bourdon, Urruty, Falou, Khalil, Shahin
  and Fernandez-Maloigne}]{alnazer2021recent}
\bibinfo{author}{Alnazer, I.}, \bibinfo{author}{Bourdon, P.},
  \bibinfo{author}{Urruty, T.}, \bibinfo{author}{Falou, O.},
  \bibinfo{author}{Khalil, M.}, \bibinfo{author}{Shahin, A.},
  \bibinfo{author}{Fernandez-Maloigne, C.}, \bibinfo{year}{2021}.
\newblock \bibinfo{title}{Recent advances in medical image processing for the
  evaluation of chronic kidney disease}.
\newblock \bibinfo{journal}{Medical Image Analysis} \bibinfo{volume}{69},
  \bibinfo{pages}{101960}.
\bibitem[{Bouteldja et~al.(2021)Bouteldja, Klinkhammer, B{\"u}low, Droste,
  Otten, von Stillfried, Moellmann, Sheehan, Korstanje, Menzel
  et~al.}]{bouteldja2021deep}
\bibinfo{author}{Bouteldja, N.}, \bibinfo{author}{Klinkhammer, B.M.},
  \bibinfo{author}{B{\"u}low, R.D.}, \bibinfo{author}{Droste, P.},
  \bibinfo{author}{Otten, S.W.}, \bibinfo{author}{von Stillfried, S.F.},
  \bibinfo{author}{Moellmann, J.}, \bibinfo{author}{Sheehan, S.M.},
  \bibinfo{author}{Korstanje, R.}, \bibinfo{author}{Menzel, S.}, et~al.,
  \bibinfo{year}{2021}.
\newblock \bibinfo{title}{Deep learning--based segmentation and quantification
  in experimental kidney histopathology}.
\newblock \bibinfo{journal}{Journal of the American Society of Nephrology:
  JASN} \bibinfo{volume}{32}, \bibinfo{pages}{52}.
\bibitem[{Brachemi and Boll{\'e}e(2014)}]{brachemi2014renal}
\bibinfo{author}{Brachemi, S.}, \bibinfo{author}{Boll{\'e}e, G.},
  \bibinfo{year}{2014}.
\newblock \bibinfo{title}{Renal biopsy practice: What is the gold standard?}
\newblock \bibinfo{journal}{World journal of nephrology} \bibinfo{volume}{3},
  \bibinfo{pages}{287}.
\bibitem[{Cai and Vasconcelos(2019)}]{cai2019cascade}
\bibinfo{author}{Cai, Z.}, \bibinfo{author}{Vasconcelos, N.},
  \bibinfo{year}{2019}.
\newblock \bibinfo{title}{Cascade r-cnn: High quality object detection and
  instance segmentation}.
\newblock \bibinfo{journal}{IEEE transactions on pattern analysis and machine
  intelligence} \bibinfo{volume}{43}, \bibinfo{pages}{1483--1498}.
\bibitem[{Carion et~al.(2020)Carion, Massa, Synnaeve, Usunier, Kirillov and
  Zagoruyko}]{carion2020end}
\bibinfo{author}{Carion, N.}, \bibinfo{author}{Massa, F.},
  \bibinfo{author}{Synnaeve, G.}, \bibinfo{author}{Usunier, N.},
  \bibinfo{author}{Kirillov, A.}, \bibinfo{author}{Zagoruyko, S.},
  \bibinfo{year}{2020}.
\newblock \bibinfo{title}{End-to-end object detection with transformers}, in:
  \bibinfo{booktitle}{European conference on computer vision},
  \bibinfo{organization}{Springer}. pp. \bibinfo{pages}{213--229}.
\bibitem[{Chen et~al.(2021)Chen, Panda and Fan}]{chen2021regionvit}
\bibinfo{author}{Chen, C.F.}, \bibinfo{author}{Panda, R.},
  \bibinfo{author}{Fan, Q.}, \bibinfo{year}{2021}.
\newblock \bibinfo{title}{Regionvit: Regional-to-local attention for vision
  transformers}.
\newblock \bibinfo{journal}{arXiv preprint arXiv:2106.02689} .
\bibitem[{Chen et~al.(2019)Chen, Wang, Pang, Cao, Xiong, Li, Sun, Feng, Liu, Xu
  et~al.}]{chen2019mmdetection}
\bibinfo{author}{Chen, K.}, \bibinfo{author}{Wang, J.}, \bibinfo{author}{Pang,
  J.}, \bibinfo{author}{Cao, Y.}, \bibinfo{author}{Xiong, Y.},
  \bibinfo{author}{Li, X.}, \bibinfo{author}{Sun, S.}, \bibinfo{author}{Feng,
  W.}, \bibinfo{author}{Liu, Z.}, \bibinfo{author}{Xu, J.}, et~al.,
  \bibinfo{year}{2019}.
\newblock \bibinfo{title}{Mmdetection: Open mmlab detection toolbox and
  benchmark}.
\newblock \bibinfo{journal}{arXiv preprint arXiv:1906.07155} .
\bibitem[{Chen et~al.(2022a)Chen, Sun, Song and Luo}]{chen2022diffusiondet}
\bibinfo{author}{Chen, S.}, \bibinfo{author}{Sun, P.}, \bibinfo{author}{Song,
  Y.}, \bibinfo{author}{Luo, P.}, \bibinfo{year}{2022}a.
\newblock \bibinfo{title}{Diffusiondet: Diffusion model for object detection}.
\newblock \bibinfo{journal}{arXiv preprint arXiv:2211.09788} .
\bibitem[{Chen et~al.(2022b)Chen, Li, Saxena, Hinton and
  Fleet}]{chen2022generalist}
\bibinfo{author}{Chen, T.}, \bibinfo{author}{Li, L.}, \bibinfo{author}{Saxena,
  S.}, \bibinfo{author}{Hinton, G.}, \bibinfo{author}{Fleet, D.J.},
  \bibinfo{year}{2022}b.
\newblock \bibinfo{title}{A generalist framework for panoptic segmentation of
  images and videos}.
\newblock \bibinfo{journal}{arXiv preprint arXiv:2210.06366} .
\bibitem[{Chen et~al.(2022c)Chen, Zhang and Hinton}]{chen2022analog}
\bibinfo{author}{Chen, T.}, \bibinfo{author}{Zhang, R.},
  \bibinfo{author}{Hinton, G.}, \bibinfo{year}{2022}c.
\newblock \bibinfo{title}{Analog bits: Generating discrete data using diffusion
  models with self-conditioning}.
\newblock \bibinfo{journal}{arXiv preprint arXiv:2208.04202} .
\bibitem[{Chen et~al.(2020)Chen, Dai, Liu, Chen, Yuan and
  Liu}]{chen2020dynamic}
\bibinfo{author}{Chen, Y.}, \bibinfo{author}{Dai, X.}, \bibinfo{author}{Liu,
  M.}, \bibinfo{author}{Chen, D.}, \bibinfo{author}{Yuan, L.},
  \bibinfo{author}{Liu, Z.}, \bibinfo{year}{2020}.
\newblock \bibinfo{title}{Dynamic convolution: Attention over convolution
  kernels}, in: \bibinfo{booktitle}{Proceedings of the IEEE/CVF conference on
  computer vision and pattern recognition}, pp. \bibinfo{pages}{11030--11039}.
\bibitem[{Cheng et~al.(2022a)Cheng, Misra, Schwing, Kirillov and
  Girdhar}]{cheng2022masked}
\bibinfo{author}{Cheng, B.}, \bibinfo{author}{Misra, I.},
  \bibinfo{author}{Schwing, A.G.}, \bibinfo{author}{Kirillov, A.},
  \bibinfo{author}{Girdhar, R.}, \bibinfo{year}{2022}a.
\newblock \bibinfo{title}{Masked-attention mask transformer for universal image
  segmentation}, in: \bibinfo{booktitle}{Proceedings of the IEEE/CVF conference
  on computer vision and pattern recognition}, pp. \bibinfo{pages}{1290--1299}.
\bibitem[{Cheng et~al.(2021)Cheng, Schwing and Kirillov}]{cheng2021per}
\bibinfo{author}{Cheng, B.}, \bibinfo{author}{Schwing, A.},
  \bibinfo{author}{Kirillov, A.}, \bibinfo{year}{2021}.
\newblock \bibinfo{title}{Per-pixel classification is not all you need for
  semantic segmentation}.
\newblock \bibinfo{journal}{Advances in Neural Information Processing Systems}
  \bibinfo{volume}{34}, \bibinfo{pages}{17864--17875}.
\bibitem[{Cheng et~al.(2022b)Cheng, Wang, Chen, Zhang, Zhang, Huang, Zhang and
  Liu}]{cheng2022sparse}
\bibinfo{author}{Cheng, T.}, \bibinfo{author}{Wang, X.}, \bibinfo{author}{Chen,
  S.}, \bibinfo{author}{Zhang, W.}, \bibinfo{author}{Zhang, Q.},
  \bibinfo{author}{Huang, C.}, \bibinfo{author}{Zhang, Z.},
  \bibinfo{author}{Liu, W.}, \bibinfo{year}{2022}b.
\newblock \bibinfo{title}{Sparse instance activation for real-time instance
  segmentation}, in: \bibinfo{booktitle}{Proceedings of the IEEE/CVF Conference
  on Computer Vision and Pattern Recognition}, pp. \bibinfo{pages}{4433--4442}.
\bibitem[{Deng et~al.(2009)Deng, Dong, Socher, Li, Li and
  Fei-Fei}]{deng2009imagenet}
\bibinfo{author}{Deng, J.}, \bibinfo{author}{Dong, W.},
  \bibinfo{author}{Socher, R.}, \bibinfo{author}{Li, L.J.},
  \bibinfo{author}{Li, K.}, \bibinfo{author}{Fei-Fei, L.},
  \bibinfo{year}{2009}.
\newblock \bibinfo{title}{Imagenet: A large-scale hierarchical image database},
  in: \bibinfo{booktitle}{2009 IEEE conference on computer vision and pattern
  recognition}, \bibinfo{organization}{Ieee}. pp. \bibinfo{pages}{248--255}.
\bibitem[{Deng et~al.(2023)Deng, Liu, Cui, Yao, Long, Asad, Womick, Zhu, Fogo,
  Zhao et~al.}]{deng2023omni}
\bibinfo{author}{Deng, R.}, \bibinfo{author}{Liu, Q.}, \bibinfo{author}{Cui,
  C.}, \bibinfo{author}{Yao, T.}, \bibinfo{author}{Long, J.},
  \bibinfo{author}{Asad, Z.}, \bibinfo{author}{Womick, R.M.},
  \bibinfo{author}{Zhu, Z.}, \bibinfo{author}{Fogo, A.B.},
  \bibinfo{author}{Zhao, S.}, et~al., \bibinfo{year}{2023}.
\newblock \bibinfo{title}{Omni-seg: A scale-aware dynamic network for renal
  pathological image segmentation}.
\newblock \bibinfo{journal}{IEEE Transactions on Biomedical Engineering} .
\bibitem[{Deng et~al.(2024)Deng, Liu, Cui, Yao, Yue, Xiong, Yu, Wu, Yin, Wang
  et~al.}]{deng2024prpseg}
\bibinfo{author}{Deng, R.}, \bibinfo{author}{Liu, Q.}, \bibinfo{author}{Cui,
  C.}, \bibinfo{author}{Yao, T.}, \bibinfo{author}{Yue, J.},
  \bibinfo{author}{Xiong, J.}, \bibinfo{author}{Yu, L.}, \bibinfo{author}{Wu,
  Y.}, \bibinfo{author}{Yin, M.}, \bibinfo{author}{Wang, Y.}, et~al.,
  \bibinfo{year}{2024}.
\newblock \bibinfo{title}{Prpseg: Universal proposition learning for panoramic
  renal pathology segmentation}, in: \bibinfo{booktitle}{Proceedings of the
  IEEE/CVF Conference on Computer Vision and Pattern Recognition}, pp.
  \bibinfo{pages}{11736--11746}.
\bibitem[{Dhariwal and Nichol(2021)}]{dhariwal2021diffusion}
\bibinfo{author}{Dhariwal, P.}, \bibinfo{author}{Nichol, A.},
  \bibinfo{year}{2021}.
\newblock \bibinfo{title}{Diffusion models beat gans on image synthesis}.
\newblock \bibinfo{journal}{Advances in neural information processing systems}
  \bibinfo{volume}{34}, \bibinfo{pages}{8780--8794}.
\bibitem[{Fan et~al.(2024)Fan, Lv, Wang, Hong, Liu, Jiang, Ni, Li and
  Pan}]{fan2024dcdiff}
\bibinfo{author}{Fan, J.}, \bibinfo{author}{Lv, T.}, \bibinfo{author}{Wang,
  P.}, \bibinfo{author}{Hong, X.}, \bibinfo{author}{Liu, Y.},
  \bibinfo{author}{Jiang, C.}, \bibinfo{author}{Ni, J.}, \bibinfo{author}{Li,
  L.}, \bibinfo{author}{Pan, X.}, \bibinfo{year}{2024}.
\newblock \bibinfo{title}{Dcdiff: Dual-granularity cooperative diffusion models
  for pathology image analysis}.
\newblock \bibinfo{journal}{IEEE Transactions on Medical Imaging} .
\bibitem[{Fang et~al.(2021)Fang, Yang, Wang, Li, Fang, Shan, Feng and
  Liu}]{fang2021instances}
\bibinfo{author}{Fang, Y.}, \bibinfo{author}{Yang, S.}, \bibinfo{author}{Wang,
  X.}, \bibinfo{author}{Li, Y.}, \bibinfo{author}{Fang, C.},
  \bibinfo{author}{Shan, Y.}, \bibinfo{author}{Feng, B.}, \bibinfo{author}{Liu,
  W.}, \bibinfo{year}{2021}.
\newblock \bibinfo{title}{Instances as queries}.
\newblock \href{http://arxiv.org/abs/2105.01928}{\tt arXiv:2105.01928}.
\bibitem[{Feng et~al.(2024)Feng, Ong, Young, Chen, Li, Huo, Lu, Gu, Liu, Tang
  et~al.}]{feng2024artificial}
\bibinfo{author}{Feng, C.}, \bibinfo{author}{Ong, K.}, \bibinfo{author}{Young,
  D.M.}, \bibinfo{author}{Chen, B.}, \bibinfo{author}{Li, L.},
  \bibinfo{author}{Huo, X.}, \bibinfo{author}{Lu, H.}, \bibinfo{author}{Gu,
  W.}, \bibinfo{author}{Liu, F.}, \bibinfo{author}{Tang, H.}, et~al.,
  \bibinfo{year}{2024}.
\newblock \bibinfo{title}{Artificial intelligence-assisted quantification and
  assessment of whole slide images for pediatric kidney disease diagnosis}.
\newblock \bibinfo{journal}{Bioinformatics} \bibinfo{volume}{40},
  \bibinfo{pages}{btad740}.
\bibitem[{Gadermayr et~al.(2019)Gadermayr, Gupta, Appel, Boor, Klinkhammer and
  Merhof}]{gadermayr2019generative}
\bibinfo{author}{Gadermayr, M.}, \bibinfo{author}{Gupta, L.},
  \bibinfo{author}{Appel, V.}, \bibinfo{author}{Boor, P.},
  \bibinfo{author}{Klinkhammer, B.M.}, \bibinfo{author}{Merhof, D.},
  \bibinfo{year}{2019}.
\newblock \bibinfo{title}{Generative adversarial networks for facilitating
  stain-independent supervised and unsupervised segmentation: a study on kidney
  histology}.
\newblock \bibinfo{journal}{IEEE transactions on medical imaging}
  \bibinfo{volume}{38}, \bibinfo{pages}{2293--2302}.
\bibitem[{Girshick(2015)}]{girshick2015fast}
\bibinfo{author}{Girshick, R.}, \bibinfo{year}{2015}.
\newblock \bibinfo{title}{Fast r-cnn}.
\newblock \href{http://arxiv.org/abs/1504.08083}{\tt arXiv:1504.08083}.
\bibitem[{Gonzalez and Woods(2006)}]{digital_image_processing}
\bibinfo{author}{Gonzalez, R.C.}, \bibinfo{author}{Woods, R.E.},
  \bibinfo{year}{2006}.
\newblock \bibinfo{title}{Digital Image Processing (3rd Edition)}.
\newblock \bibinfo{publisher}{Prentice-Hall, Inc.}, \bibinfo{address}{USA}.
\bibitem[{Gu et~al.(2022)Gu, Chen, Xu, Lan, Meng and
  Wang}]{gu2022diffusioninst}
\bibinfo{author}{Gu, Z.}, \bibinfo{author}{Chen, H.}, \bibinfo{author}{Xu, Z.},
  \bibinfo{author}{Lan, J.}, \bibinfo{author}{Meng, C.}, \bibinfo{author}{Wang,
  W.}, \bibinfo{year}{2022}.
\newblock \bibinfo{title}{Diffusioninst: Diffusion model for instance
  segmentation}.
\newblock \bibinfo{journal}{arXiv preprint arXiv:2212.02773} .
\bibitem[{He et~al.(2017)He, Gkioxari, Doll{\'a}r and Girshick}]{he2017mask}
\bibinfo{author}{He, K.}, \bibinfo{author}{Gkioxari, G.},
  \bibinfo{author}{Doll{\'a}r, P.}, \bibinfo{author}{Girshick, R.},
  \bibinfo{year}{2017}.
\newblock \bibinfo{title}{Mask r-cnn}, in: \bibinfo{booktitle}{Proceedings of
  the IEEE international conference on computer vision}, pp.
  \bibinfo{pages}{2961--2969}.
\bibitem[{He et~al.(2015)He, Zhang, Ren and Sun}]{he2015deep}
\bibinfo{author}{He, K.}, \bibinfo{author}{Zhang, X.}, \bibinfo{author}{Ren,
  S.}, \bibinfo{author}{Sun, J.}, \bibinfo{year}{2015}.
\newblock \bibinfo{title}{Deep residual learning for image recognition}.
\newblock \href{http://arxiv.org/abs/1512.03385}{\tt arXiv:1512.03385}.
\bibitem[{Hermsen et~al.(2019)Hermsen, de~Bel, Den~Boer, Steenbergen, Kers,
  Florquin, Roelofs, Stegall, Alexander, Smith et~al.}]{hermsen2019deep}
\bibinfo{author}{Hermsen, M.}, \bibinfo{author}{de~Bel, T.},
  \bibinfo{author}{Den~Boer, M.}, \bibinfo{author}{Steenbergen, E.J.},
  \bibinfo{author}{Kers, J.}, \bibinfo{author}{Florquin, S.},
  \bibinfo{author}{Roelofs, J.J.}, \bibinfo{author}{Stegall, M.D.},
  \bibinfo{author}{Alexander, M.P.}, \bibinfo{author}{Smith, B.H.}, et~al.,
  \bibinfo{year}{2019}.
\newblock \bibinfo{title}{Deep learning--based histopathologic assessment of
  kidney tissue}.
\newblock \bibinfo{journal}{Journal of the American Society of Nephrology:
  JASN} \bibinfo{volume}{30}, \bibinfo{pages}{1968}.
\bibitem[{Ho et~al.(2020)Ho, Jain and Abbeel}]{ho2020denoising}
\bibinfo{author}{Ho, J.}, \bibinfo{author}{Jain, A.}, \bibinfo{author}{Abbeel,
  P.}, \bibinfo{year}{2020}.
\newblock \bibinfo{title}{Denoising diffusion probabilistic models}.
\newblock \bibinfo{journal}{Advances in neural information processing systems}
  \bibinfo{volume}{33}, \bibinfo{pages}{6840--6851}.
\bibitem[{Huang and Zambrini(2023)}]{huang2023stochastic}
\bibinfo{author}{Huang, Q.}, \bibinfo{author}{Zambrini, J.C.},
  \bibinfo{year}{2023}.
\newblock \bibinfo{title}{Stochastic geometric mechanics in nonequilibrium
  thermodynamics: Schr{\"o}dinger meets onsager}.
\newblock \bibinfo{journal}{Journal of Physics A: Mathematical and Theoretical}
  \bibinfo{volume}{56}, \bibinfo{pages}{134003}.
\bibitem[{Huang et~al.(2019)Huang, Huang, Gong, Huang and Wang}]{huang2019mask}
\bibinfo{author}{Huang, Z.}, \bibinfo{author}{Huang, L.},
  \bibinfo{author}{Gong, Y.}, \bibinfo{author}{Huang, C.},
  \bibinfo{author}{Wang, X.}, \bibinfo{year}{2019}.
\newblock \bibinfo{title}{Mask scoring r-cnn}.
\newblock \href{http://arxiv.org/abs/1903.00241}{\tt arXiv:1903.00241}.
\bibitem[{Jha et~al.(2021)Jha, Yang, Deng, Kapp, Fogo and
  Huo}]{jha2021instance}
\bibinfo{author}{Jha, A.}, \bibinfo{author}{Yang, H.}, \bibinfo{author}{Deng,
  R.}, \bibinfo{author}{Kapp, M.E.}, \bibinfo{author}{Fogo, A.B.},
  \bibinfo{author}{Huo, Y.}, \bibinfo{year}{2021}.
\newblock \bibinfo{title}{Instance segmentation for whole slide imaging:
  end-to-end or detect-then-segment}.
\newblock \bibinfo{journal}{Journal of Medical Imaging} \bibinfo{volume}{8},
  \bibinfo{pages}{014001--014001}.
\bibitem[{Jiang et~al.(2021)Jiang, Chen, Dong, Mei, Zhu, Liu, Cai, Yan, Wang,
  Zuo et~al.}]{jiang2021deep}
\bibinfo{author}{Jiang, L.}, \bibinfo{author}{Chen, W.}, \bibinfo{author}{Dong,
  B.}, \bibinfo{author}{Mei, K.}, \bibinfo{author}{Zhu, C.},
  \bibinfo{author}{Liu, J.}, \bibinfo{author}{Cai, M.}, \bibinfo{author}{Yan,
  Y.}, \bibinfo{author}{Wang, G.}, \bibinfo{author}{Zuo, L.}, et~al.,
  \bibinfo{year}{2021}.
\newblock \bibinfo{title}{A deep learning-based approach for glomeruli instance
  segmentation from multistained renal biopsy pathologic images}.
\newblock \bibinfo{journal}{The American Journal of Pathology}
  \bibinfo{volume}{191}, \bibinfo{pages}{1431--1441}.
\bibitem[{Kang et~al.(2021)Kang, Luo, Feng, Zeng, Quan, Hu and
  Liu}]{kang2021stainnet}
\bibinfo{author}{Kang, H.}, \bibinfo{author}{Luo, D.}, \bibinfo{author}{Feng,
  W.}, \bibinfo{author}{Zeng, S.}, \bibinfo{author}{Quan, T.},
  \bibinfo{author}{Hu, J.}, \bibinfo{author}{Liu, X.}, \bibinfo{year}{2021}.
\newblock \bibinfo{title}{Stainnet: a fast and robust stain normalization
  network}.
\newblock \bibinfo{journal}{Frontiers in Medicine} \bibinfo{volume}{8},
  \bibinfo{pages}{746307}.
\bibitem[{Li et~al.(2022a)Li, Zhang, Liu, Guo, Ni and Zhang}]{li2022dn}
\bibinfo{author}{Li, F.}, \bibinfo{author}{Zhang, H.}, \bibinfo{author}{Liu,
  S.}, \bibinfo{author}{Guo, J.}, \bibinfo{author}{Ni, L.M.},
  \bibinfo{author}{Zhang, L.}, \bibinfo{year}{2022}a.
\newblock \bibinfo{title}{Dn-detr: Accelerate detr training by introducing
  query denoising}, in: \bibinfo{booktitle}{Proceedings of the IEEE/CVF
  Conference on Computer Vision and Pattern Recognition}, pp.
  \bibinfo{pages}{13619--13627}.
\bibitem[{Li et~al.(2023)Li, Zhang, Xu, Liu, Zhang, Ni and Shum}]{li2023mask}
\bibinfo{author}{Li, F.}, \bibinfo{author}{Zhang, H.}, \bibinfo{author}{Xu,
  H.}, \bibinfo{author}{Liu, S.}, \bibinfo{author}{Zhang, L.},
  \bibinfo{author}{Ni, L.M.}, \bibinfo{author}{Shum, H.Y.},
  \bibinfo{year}{2023}.
\newblock \bibinfo{title}{Mask dino: Towards a unified transformer-based
  framework for object detection and segmentation}, in:
  \bibinfo{booktitle}{Proceedings of the IEEE/CVF Conference on Computer Vision
  and Pattern Recognition}, pp. \bibinfo{pages}{3041--3050}.
\bibitem[{Li et~al.(2024)Li, Zheng, Zhu, Sun, Chen, Shui, Zhang, Li and
  Yang}]{li2024pathup}
\bibinfo{author}{Li, J.}, \bibinfo{author}{Zheng, S.}, \bibinfo{author}{Zhu,
  C.}, \bibinfo{author}{Sun, Y.}, \bibinfo{author}{Chen, P.},
  \bibinfo{author}{Shui, Z.}, \bibinfo{author}{Zhang, Y.}, \bibinfo{author}{Li,
  H.}, \bibinfo{author}{Yang, L.}, \bibinfo{year}{2024}.
\newblock \bibinfo{title}{Pathup: Patch-wise timestep tracking for multi-class
  large pathology image synthesising diffusion model}, in:
  \bibinfo{booktitle}{Proceedings of the 32nd ACM International Conference on
  Multimedia}, pp. \bibinfo{pages}{3984--3993}.
\bibitem[{Li et~al.(2022b)Li, Thickstun, Gulrajani, Liang and
  Hashimoto}]{li2022diffusion}
\bibinfo{author}{Li, X.}, \bibinfo{author}{Thickstun, J.},
  \bibinfo{author}{Gulrajani, I.}, \bibinfo{author}{Liang, P.S.},
  \bibinfo{author}{Hashimoto, T.B.}, \bibinfo{year}{2022}b.
\newblock \bibinfo{title}{Diffusion-lm improves controllable text generation}.
\newblock \bibinfo{journal}{Advances in Neural Information Processing Systems}
  \bibinfo{volume}{35}, \bibinfo{pages}{4328--4343}.
\bibitem[{Lin et~al.(2023)Lin, Zhang, Long, Zhang, Lu, Geng, Zhou, Feng, Lu and
  Cao}]{lin2023gclr}
\bibinfo{author}{Lin, G.}, \bibinfo{author}{Zhang, Z.}, \bibinfo{author}{Long,
  K.}, \bibinfo{author}{Zhang, Y.}, \bibinfo{author}{Lu, Y.},
  \bibinfo{author}{Geng, J.}, \bibinfo{author}{Zhou, Z.},
  \bibinfo{author}{Feng, Q.}, \bibinfo{author}{Lu, L.}, \bibinfo{author}{Cao,
  L.}, \bibinfo{year}{2023}.
\newblock \bibinfo{title}{Gclr: A self-supervised representation learning
  pretext task for glomerular filtration barrier segmentation in tem images}.
\newblock \bibinfo{journal}{Artificial Intelligence in Medicine}
  \bibinfo{volume}{146}, \bibinfo{pages}{102720}.
\bibitem[{Lin et~al.(2017)Lin, Dollár, Girshick, He, Hariharan and
  Belongie}]{lin2017feature}
\bibinfo{author}{Lin, T.Y.}, \bibinfo{author}{Dollár, P.},
  \bibinfo{author}{Girshick, R.}, \bibinfo{author}{He, K.},
  \bibinfo{author}{Hariharan, B.}, \bibinfo{author}{Belongie, S.},
  \bibinfo{year}{2017}.
\newblock \bibinfo{title}{Feature pyramid networks for object detection}.
\newblock \href{http://arxiv.org/abs/1612.03144}{\tt arXiv:1612.03144}.
\bibitem[{Lin et~al.(2014)Lin, Maire, Belongie, Hays, Perona, Ramanan,
  Doll{\'a}r and Zitnick}]{lin2014microsoft}
\bibinfo{author}{Lin, T.Y.}, \bibinfo{author}{Maire, M.},
  \bibinfo{author}{Belongie, S.}, \bibinfo{author}{Hays, J.},
  \bibinfo{author}{Perona, P.}, \bibinfo{author}{Ramanan, D.},
  \bibinfo{author}{Doll{\'a}r, P.}, \bibinfo{author}{Zitnick, C.L.},
  \bibinfo{year}{2014}.
\newblock \bibinfo{title}{Microsoft coco: Common objects in context}, in:
  \bibinfo{booktitle}{Computer Vision--ECCV 2014: 13th European Conference,
  Zurich, Switzerland, September 6-12, 2014, Proceedings, Part V 13},
  \bibinfo{organization}{Springer}. pp. \bibinfo{pages}{740--755}.
\bibitem[{Liu et~al.(2023)Liu, Wu, Chen, Hui, Zhang, Hao, Lu, Cheng, Zeng, Han
  et~al.}]{liu2023diagnosis}
\bibinfo{author}{Liu, X.}, \bibinfo{author}{Wu, Y.}, \bibinfo{author}{Chen,
  Y.}, \bibinfo{author}{Hui, D.}, \bibinfo{author}{Zhang, J.},
  \bibinfo{author}{Hao, F.}, \bibinfo{author}{Lu, Y.}, \bibinfo{author}{Cheng,
  H.}, \bibinfo{author}{Zeng, Y.}, \bibinfo{author}{Han, W.}, et~al.,
  \bibinfo{year}{2023}.
\newblock \bibinfo{title}{Diagnosis of diabetic kidney disease in whole slide
  images via ai-driven quantification of pathological indicators}.
\newblock \bibinfo{journal}{Computers in Biology and Medicine}
  \bibinfo{volume}{166}, \bibinfo{pages}{107470}.
\bibitem[{Meseguer et~al.(2024)Meseguer, Del~Amor and
  Naranjo}]{meseguer2024micil}
\bibinfo{author}{Meseguer, P.}, \bibinfo{author}{Del~Amor, R.},
  \bibinfo{author}{Naranjo, V.}, \bibinfo{year}{2024}.
\newblock \bibinfo{title}{Micil: Multiple-instance class-incremental learning
  for skin cancer whole slide images}.
\newblock \bibinfo{journal}{Artificial Intelligence in Medicine}
  \bibinfo{volume}{152}, \bibinfo{pages}{102870}.
\bibitem[{Mohamed and Lakshminarayanan(2016)}]{mohamed2016learning}
\bibinfo{author}{Mohamed, S.}, \bibinfo{author}{Lakshminarayanan, B.},
  \bibinfo{year}{2016}.
\newblock \bibinfo{title}{Learning in implicit generative models}.
\newblock \bibinfo{journal}{arXiv preprint arXiv:1610.03483} .
\bibitem[{Nichol and Dhariwal(2021)}]{nichol2021improved}
\bibinfo{author}{Nichol, A.Q.}, \bibinfo{author}{Dhariwal, P.},
  \bibinfo{year}{2021}.
\newblock \bibinfo{title}{Improved denoising diffusion probabilistic models},
  in: \bibinfo{booktitle}{International Conference on Machine Learning},
  \bibinfo{organization}{PMLR}. pp. \bibinfo{pages}{8162--8171}.
\bibitem[{Oh and Jeong(2023)}]{oh2023diffmix}
\bibinfo{author}{Oh, H.J.}, \bibinfo{author}{Jeong, W.K.},
  \bibinfo{year}{2023}.
\newblock \bibinfo{title}{Diffmix: Diffusion model-based data synthesis for
  nuclei segmentation and classification in imbalanced pathology image
  datasets}, in: \bibinfo{booktitle}{International Conference on Medical Image
  Computing and Computer-Assisted Intervention},
  \bibinfo{organization}{Springer}. pp. \bibinfo{pages}{337--345}.
\bibitem[{Popov et~al.(2021)Popov, Vovk, Gogoryan, Sadekova and
  Kudinov}]{popov2021gradtts}
\bibinfo{author}{Popov, V.}, \bibinfo{author}{Vovk, I.},
  \bibinfo{author}{Gogoryan, V.}, \bibinfo{author}{Sadekova, T.},
  \bibinfo{author}{Kudinov, M.}, \bibinfo{year}{2021}.
\newblock \bibinfo{title}{Grad-tts: A diffusion probabilistic model for
  text-to-speech}.
\newblock \href{http://arxiv.org/abs/2105.06337}{\tt arXiv:2105.06337}.
\bibitem[{Ren et~al.(2016)Ren, He, Girshick and Sun}]{ren2016faster}
\bibinfo{author}{Ren, S.}, \bibinfo{author}{He, K.}, \bibinfo{author}{Girshick,
  R.}, \bibinfo{author}{Sun, J.}, \bibinfo{year}{2016}.
\newblock \bibinfo{title}{Faster r-cnn: Towards real-time object detection with
  region proposal networks}.
\newblock \href{http://arxiv.org/abs/1506.01497}{\tt arXiv:1506.01497}.
\bibitem[{Rombach et~al.(2022)Rombach, Blattmann, Lorenz, Esser and
  Ommer}]{rombach2022high}
\bibinfo{author}{Rombach, R.}, \bibinfo{author}{Blattmann, A.},
  \bibinfo{author}{Lorenz, D.}, \bibinfo{author}{Esser, P.},
  \bibinfo{author}{Ommer, B.}, \bibinfo{year}{2022}.
\newblock \bibinfo{title}{High-resolution image synthesis with latent diffusion
  models}, in: \bibinfo{booktitle}{Proceedings of the IEEE/CVF conference on
  computer vision and pattern recognition}, pp. \bibinfo{pages}{10684--10695}.
\bibitem[{Salvi et~al.(2021)Salvi, Mogetta, Gambella, Molinaro, Barreca,
  Papotti and Molinari}]{salvi2021automated}
\bibinfo{author}{Salvi, M.}, \bibinfo{author}{Mogetta, A.},
  \bibinfo{author}{Gambella, A.}, \bibinfo{author}{Molinaro, L.},
  \bibinfo{author}{Barreca, A.}, \bibinfo{author}{Papotti, M.},
  \bibinfo{author}{Molinari, F.}, \bibinfo{year}{2021}.
\newblock \bibinfo{title}{Automated assessment of glomerulosclerosis and
  tubular atrophy using deep learning}.
\newblock \bibinfo{journal}{Computerized Medical Imaging and Graphics}
  \bibinfo{volume}{90}, \bibinfo{pages}{101930}.
\bibitem[{Shickel et~al.(2023)Shickel, Lucarelli, Rao, Yun, Moon, Seok and
  Sarder}]{shickel2023spatially}
\bibinfo{author}{Shickel, B.}, \bibinfo{author}{Lucarelli, N.},
  \bibinfo{author}{Rao, A.}, \bibinfo{author}{Yun, D.}, \bibinfo{author}{Moon,
  K.C.}, \bibinfo{author}{Seok, H.S.}, \bibinfo{author}{Sarder, P.},
  \bibinfo{year}{2023}.
\newblock \bibinfo{title}{Spatially aware transformer networks for contextual
  prediction of diabetic nephropathy progression from whole slide images}, in:
  \bibinfo{booktitle}{Medical Imaging 2023: Digital and Computational
  Pathology}, \bibinfo{organization}{SPIE}. pp. \bibinfo{pages}{129--140}.
\bibitem[{Sohl-Dickstein et~al.(2015)Sohl-Dickstein, Weiss, Maheswaranathan and
  Ganguli}]{sohl2015deep}
\bibinfo{author}{Sohl-Dickstein, J.}, \bibinfo{author}{Weiss, E.},
  \bibinfo{author}{Maheswaranathan, N.}, \bibinfo{author}{Ganguli, S.},
  \bibinfo{year}{2015}.
\newblock \bibinfo{title}{Deep unsupervised learning using nonequilibrium
  thermodynamics}, in: \bibinfo{booktitle}{International conference on machine
  learning}, \bibinfo{organization}{PMLR}. pp. \bibinfo{pages}{2256--2265}.
\bibitem[{Song et~al.(2020)Song, Meng and Ermon}]{song2020denoising}
\bibinfo{author}{Song, J.}, \bibinfo{author}{Meng, C.}, \bibinfo{author}{Ermon,
  S.}, \bibinfo{year}{2020}.
\newblock \bibinfo{title}{Denoising diffusion implicit models}.
\newblock \bibinfo{journal}{arXiv preprint arXiv:2010.02502} .
\bibitem[{Srinidhi et~al.(2021)Srinidhi, Ciga and Martel}]{srinidhi2021deep}
\bibinfo{author}{Srinidhi, C.L.}, \bibinfo{author}{Ciga, O.},
  \bibinfo{author}{Martel, A.L.}, \bibinfo{year}{2021}.
\newblock \bibinfo{title}{Deep neural network models for computational
  histopathology: A survey}.
\newblock \bibinfo{journal}{Medical image analysis} \bibinfo{volume}{67},
  \bibinfo{pages}{101813}.
\bibitem[{Sun et~al.(2021)Sun, Jiang, Xie, Shao, Yuan, Wang and
  Luo}]{sun2021makes}
\bibinfo{author}{Sun, P.}, \bibinfo{author}{Jiang, Y.}, \bibinfo{author}{Xie,
  E.}, \bibinfo{author}{Shao, W.}, \bibinfo{author}{Yuan, Z.},
  \bibinfo{author}{Wang, C.}, \bibinfo{author}{Luo, P.}, \bibinfo{year}{2021}.
\newblock \bibinfo{title}{What makes for end-to-end object detection?}, in:
  \bibinfo{booktitle}{International Conference on Machine Learning},
  \bibinfo{organization}{PMLR}. pp. \bibinfo{pages}{9934--9944}.
\bibitem[{Vasiljevi{\'c} et~al.(2021)Vasiljevi{\'c}, Feuerhake, Wemmert and
  Lampert}]{vasiljevic2021towards}
\bibinfo{author}{Vasiljevi{\'c}, J.}, \bibinfo{author}{Feuerhake, F.},
  \bibinfo{author}{Wemmert, C.}, \bibinfo{author}{Lampert, T.},
  \bibinfo{year}{2021}.
\newblock \bibinfo{title}{Towards histopathological stain invariance by
  unsupervised domain augmentation using generative adversarial networks}.
\newblock \bibinfo{journal}{Neurocomputing} \bibinfo{volume}{460},
  \bibinfo{pages}{277--291}.
\bibitem[{Vaswani et~al.(2017)Vaswani, Shazeer, Parmar, Uszkoreit, Jones,
  Gomez, Kaiser and Polosukhin}]{vaswani2017attention}
\bibinfo{author}{Vaswani, A.}, \bibinfo{author}{Shazeer, N.},
  \bibinfo{author}{Parmar, N.}, \bibinfo{author}{Uszkoreit, J.},
  \bibinfo{author}{Jones, L.}, \bibinfo{author}{Gomez, A.N.},
  \bibinfo{author}{Kaiser, {\L}.}, \bibinfo{author}{Polosukhin, I.},
  \bibinfo{year}{2017}.
\newblock \bibinfo{title}{Attention is all you need}.
\newblock \bibinfo{journal}{Advances in neural information processing systems}
  \bibinfo{volume}{30}.
\bibitem[{Xu et~al.(2014)Xu, Mo, Feng, Zhong, Lai, Eric and Chang}]{xu2014deep}
\bibinfo{author}{Xu, Y.}, \bibinfo{author}{Mo, T.}, \bibinfo{author}{Feng, Q.},
  \bibinfo{author}{Zhong, P.}, \bibinfo{author}{Lai, M.},
  \bibinfo{author}{Eric, I.}, \bibinfo{author}{Chang, C.},
  \bibinfo{year}{2014}.
\newblock \bibinfo{title}{Deep learning of feature representation with multiple
  instance learning for medical image analysis}, in: \bibinfo{booktitle}{2014
  IEEE international conference on acoustics, speech and signal processing
  (ICASSP)}, \bibinfo{organization}{IEEE}. pp. \bibinfo{pages}{1626--1630}.
\bibitem[{Yuan et~al.(2023)Yuan, Xia, Dong, Chen, Yao, Qiu, Yan, Yin, Shi, Chen
  et~al.}]{yuan2023devil}
\bibinfo{author}{Yuan, M.}, \bibinfo{author}{Xia, Y.}, \bibinfo{author}{Dong,
  H.}, \bibinfo{author}{Chen, Z.}, \bibinfo{author}{Yao, J.},
  \bibinfo{author}{Qiu, M.}, \bibinfo{author}{Yan, K.}, \bibinfo{author}{Yin,
  X.}, \bibinfo{author}{Shi, Y.}, \bibinfo{author}{Chen, X.}, et~al.,
  \bibinfo{year}{2023}.
\newblock \bibinfo{title}{Devil is in the queries: advancing mask transformers
  for real-world medical image segmentation and out-of-distribution
  localization}, in: \bibinfo{booktitle}{Proceedings of the IEEE/CVF Conference
  on Computer Vision and Pattern Recognition}, pp.
  \bibinfo{pages}{23879--23889}.
\bibitem[{Zhang et~al.(2022)Zhang, Li, Liu, Zhang, Su, Zhu, Ni and
  Shum}]{zhang2022dino}
\bibinfo{author}{Zhang, H.}, \bibinfo{author}{Li, F.}, \bibinfo{author}{Liu,
  S.}, \bibinfo{author}{Zhang, L.}, \bibinfo{author}{Su, H.},
  \bibinfo{author}{Zhu, J.}, \bibinfo{author}{Ni, L.M.}, \bibinfo{author}{Shum,
  H.Y.}, \bibinfo{year}{2022}.
\newblock \bibinfo{title}{Dino: Detr with improved denoising anchor boxes for
  end-to-end object detection}.
\newblock \bibinfo{journal}{arXiv preprint arXiv:2203.03605} .
\bibitem[{Zhang et~al.(2023)Zhang, Li, Chen, Yuille, Liu and
  Zhou}]{zhang2023continual}
\bibinfo{author}{Zhang, Y.}, \bibinfo{author}{Li, X.}, \bibinfo{author}{Chen,
  H.}, \bibinfo{author}{Yuille, A.L.}, \bibinfo{author}{Liu, Y.},
  \bibinfo{author}{Zhou, Z.}, \bibinfo{year}{2023}.
\newblock \bibinfo{title}{Continual learning for abdominal multi-organ and
  tumor segmentation}, in: \bibinfo{booktitle}{International conference on
  medical image computing and computer-assisted intervention},
  \bibinfo{organization}{Springer}. pp. \bibinfo{pages}{35--45}.
\bibitem[{Zhu et~al.(2020)Zhu, Su, Lu, Li, Wang and Dai}]{zhu2020deformable}
\bibinfo{author}{Zhu, X.}, \bibinfo{author}{Su, W.}, \bibinfo{author}{Lu, L.},
  \bibinfo{author}{Li, B.}, \bibinfo{author}{Wang, X.}, \bibinfo{author}{Dai,
  J.}, \bibinfo{year}{2020}.
\newblock \bibinfo{title}{Deformable detr: Deformable transformers for
  end-to-end object detection}.
\newblock \bibinfo{journal}{arXiv preprint arXiv:2010.04159} .

\end{thebibliography}
\pagebreak
\appendix
\addcontentsline{toc}{section}{Appendices}
{
\section{Additional Visualization Samples}
As an extra, we have added two zoom-in samples to highlight the details of our method and other approaches. The first row of Figure A.17 shows that our method can correctly divide two glomeruli without overlap. Cascade Mask RCNN and Mask RCNN, however, generate multiple responses at the border of two objects. QueryInst with static queries cannot detect and segment any glomerulus. The last row of Figure A.17 demonstrates four large arteries in a line. It is complicated since two large arteries pass over most input image regions. Although it had slightly multiple responses at the artery border, our method can capture long-range spatial dependencies for large objects using attention mechanisms with dynamic queries. In contrast, Cascade Mask RCNN and Mask RCNN cannot model large objects without attention modules.  QueryInst adopts an attention mechanism with static queries and cannot process complicated combinations of shapes and appearances with fixed query numbers.}  
\begin{figure*}[!thbp]
  \centering
  {\includegraphics[width=0.95\textwidth, keepaspectratio=true]{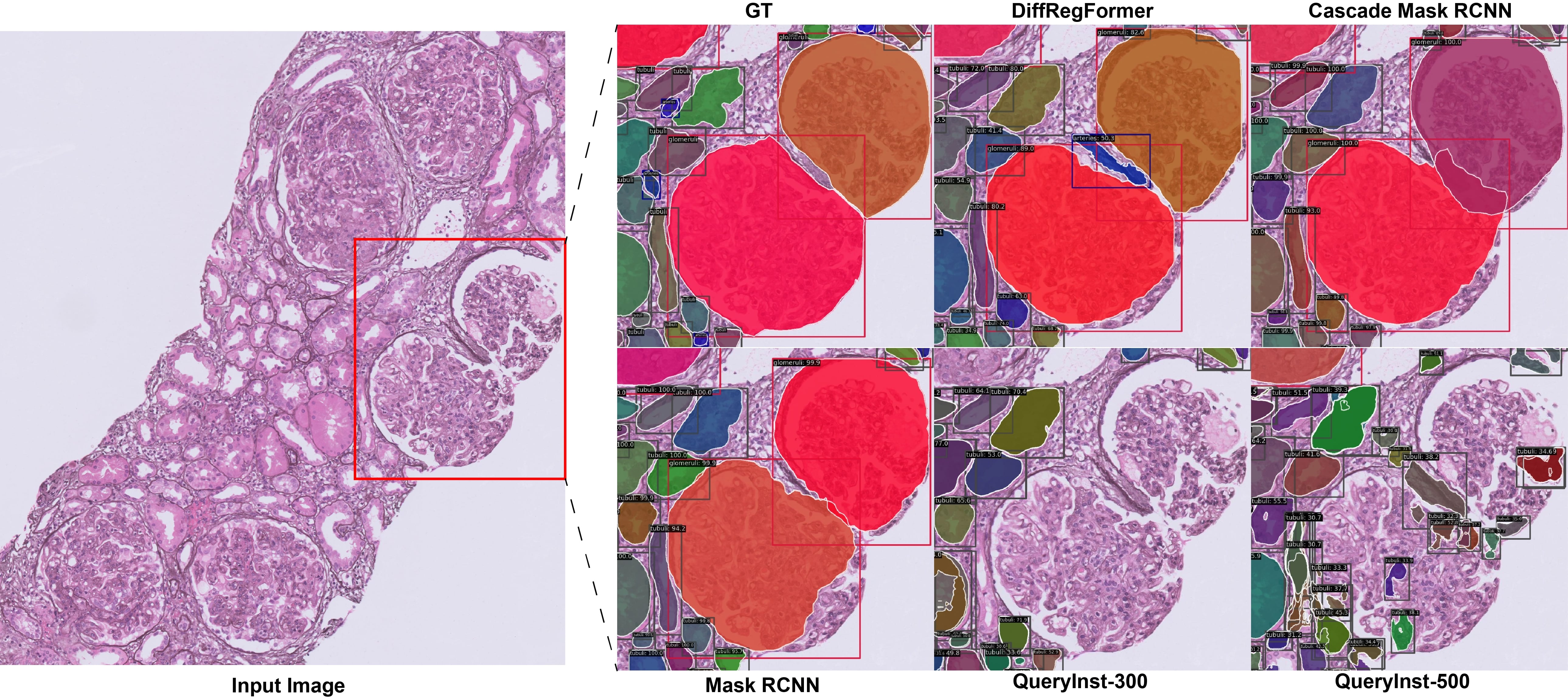}}
  \vspace{3mm}
  \makebox[0.8\textwidth]{\includegraphics[width=0.95\textwidth, keepaspectratio=true]{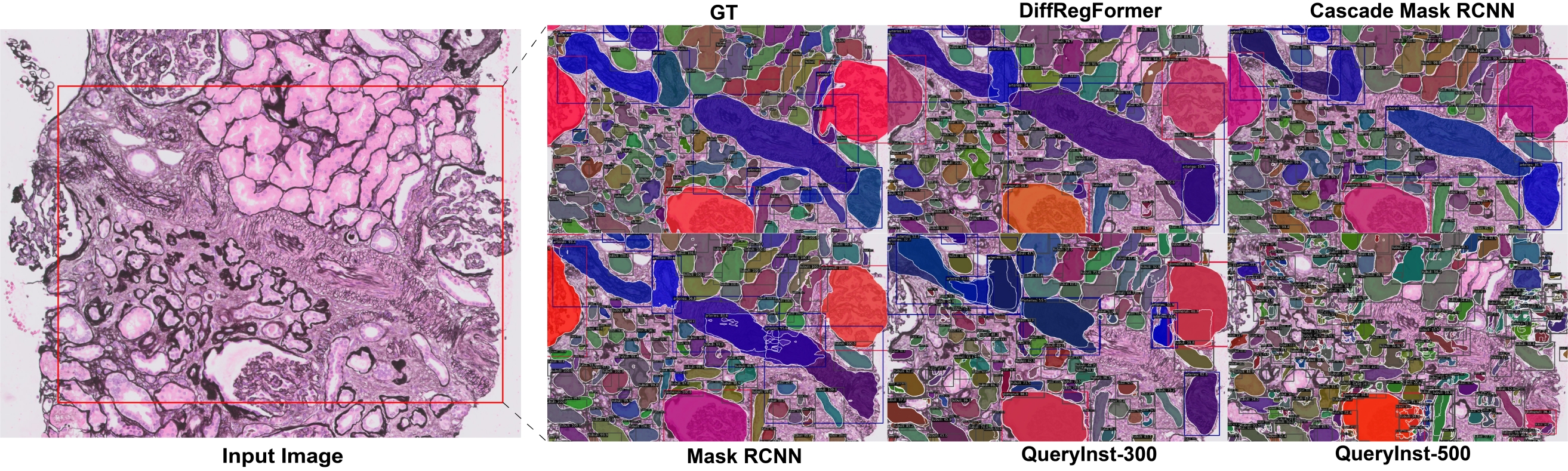}}
    \label{fig:vis_zoom_in}
    \caption{Detailed zoom-in samples comparing DiffRegFormer with other methods.}
\end{figure*}


\end{document}